\documentclass[sn-mathphys-num]{sn-jnl}

\usepackage{graphicx}%
\usepackage{multirow}%
\usepackage{amsmath,amssymb,amsfonts}%
\usepackage{amsthm}%
\usepackage{mathrsfs}%
\usepackage[title]{appendix}%
\usepackage[table]{xcolor}%
\usepackage{textcomp}%
\usepackage{manyfoot}%
\usepackage{booktabs}%
\usepackage{algorithm}%
\usepackage{algorithmicx}%
\usepackage{algpseudocode}%
\usepackage{listings}%
\usepackage{multirow}%
\usepackage{nicematrix}%
\usepackage{adjustbox}%
\usepackage{booktabs}%
\usepackage{hhline}
\usepackage{subcaption}

\theoremstyle{thmstyleone}%
\theoremstyle{thmstyletwo}%

\theoremstyle{thmstylethree}%

\begin{document}

\title[Article Title]{Transfer or Self-Supervised? Bridging the Performance Gap in Medical Imaging}


\author[1]{\fnm{Zehui} \sur{ Zhao}}\email{zehui.zhao@hdr.qut.edu.au}

\author*[2]{\fnm{Laith} \sur{Alzubaidi}}\email{l.alzubaidi@qut.edu.au}

\author[1]{\fnm{Jinglan} \sur{Zhang}}\email{jinglan.zhang@qut.edu.au}

\author[3]{\fnm{Ye} \sur{Duan}}\email{duan@clemson.edu}

\author[4]{\fnm{Usman} \sur{Naseem}}\email{usman.naseem@mq.edu.au}

\author[2]{\fnm{Yuantong} \sur{Gu}}\email{yuantong.gu@qut.edu.au}

\affil[1]{\orgdiv{School of Computer Science}, \orgname{Queensland University of Technology}, \orgaddress{\street{2 George Street}, \city{Brisbane}, \postcode{4000}, \state{Queensland}, \country{Australia}}}

\affil*[2]{\orgdiv{School of Mechanical, Medical, and Process Engineering}, \orgname{Queensland University of Technology}, \orgaddress{\street{2 George Street}, \city{Brisbane}, \postcode{4000}, \state{Queensland}, \country{Australia}}}

\affil[3]{\orgdiv{School of Computing}, \orgname{Clemson University}, \orgaddress{ \city{Clemson}, \postcode{29631}, \state{South Carolina}, \country{USA}}}

\affil[4]{\orgdiv{School of Computing}, \orgname{Macquarie University}, \orgaddress{ \city{Sydney}, \postcode{2109}, \state{New South Wales}, \country{Australia}}}


\abstract{
\textbf{Background}

Recently, transfer learning and self-supervised learning have gained significant attention within the medical field due to their ability to mitigate the challenges posed by limited data availability, improve model generalisation, and reduce computational expenses. Leveraging pre-trained features from analogous domains, these methodologies prove instrumental in improving model performance and addressing data scarcity issues stemming from the privacy concerns associated with patient data sharing and the exorbitant costs involved in collecting medical test data. Consequently, transfer learning and self-supervised learning hold immense potential for advancing medical research. However, recent work noticed a performance gap between transfer and self-supervised learning with different dataset sizes and application areas, especially in the medical field. Asserting the superiority of one method over the other across all contexts is irresponsible, leaving researchers grappling with the challenge of selecting the most fitting method and configuration for their specific scenarios. This study carried out a comparative study between transfer learning and self-supervised learning methods through experimentation to examine the effect of data characteristics on these methods. Moreover, we proposed a new approach to address this performance gap in medical environments. 

\textbf{Methods}

Specifically, multiple widely used convolutional neural network models (ResNet, MobileNet, InceptionNet, and Xception) were employed to pre-trained under transfer learning and self-supervised learning settings respectively. These pre-trained models were then tested using four different datasets from medical domains, including colourful and grey-scale image samples. Based on the performance gap that transfer learning and self-supervised learning show, we developed a new approach utilising both deep convolutional generative adversarial networks and double fine-tuning techniques to enhance the performance and robustness of transfer learning and self-supervised learning techniques. 

\textbf{Results}

The findings of this study revealed that the transfer learning setting demonstrated better performance on colourful and large-size medical datasets. In contrast, the self-supervised learning method has shown superior performance on grey-scale datasets and is more robust towards limited data scenarios. Furthermore, the data imbalance problem and domain mismatch issue that occurs frequently in the medical field also have a significant bad effect on model performance and robustness. The experimental result of the proposed approach has been proven to effectively reduce the aforementioned problems without needing a large number of pre-training data, achieving a superior accuracy performance of 90.67\%, 97.22\%, 96.40\%, and 92.64\% among multiple challenging medical datasets and outperforming the state-of-the-art works.

\textbf{Conclusions}

This comprehensive analysis stresses the influence of data imbalance, dataset size, and dissimilar transfer issues in the medical field, and we noticed that the difference in colour information between the source domain and the target domain may enlarge the model performance gap between the transfer learning and self-supervised learning paradigm. The proposed approach does not just improve the model's performance towards the aforementioned challenging scenarios but also improves the trained model's robustness, suggesting a potential way to migrate the performance gap between two pre-training paradigms. The insights gleaned from this research provide valuable guidance in selecting the appropriate pre-training settings in various scenarios, with potential applications extending beyond medical research.

\textbf{Trial registration}

Clinical trial number: not applicable.

}




\maketitle

\section{Introduction}\label{sec1}

In recent years, Deep Learning (DL) has emerged as a pivotal contributor to advances in computer vision, particularly within the medical domain. The significance of DL lies in training high-performance models, while significantly boosting medical automation and reducing the use of human resources in the clinical environment. However, traditional supervised DL architectures, exemplified by models such as AlexNet \citep{krizhevsky2012imagenet} and ResNet \citep{he2016deep}, are based heavily on abundant, well-labelled data for feature acquisition and subsequent task support. In the medical realm, where data annotation requires expert intervention, the scarcity and expense of labelled data make the application of the DL technique increasingly difficult. 

To tackle this issue, Transfer Learning (TL) has been proposed to produce a simple way for reusing existing knowledge and enforcing the capability of trained models \citep{niu2020decade}. On the other hand, Self-Supervised Learning (SSL), a more recent entrant, has gained prominence by achieving state-of-the-art supervised representation learning model performance without relying on additional manual annotations. Recognising the practical importance of TL and SSL in computer vision tasks, researchers have diligently benchmarked their performance, seeking information for future architectural advancements \citep{hosseinzadeh2021systematic}. Recent investigations encompass diverse aspects such as the suitability of different architectures, the relevance of source and target domains, the exploration of various pretext tasks, analysis of the impact of the size of the pre-training dataset, and evaluation of the model capacity on the performance of downstream tasks \citep{neyshabur2020being, ericsson2022self2}. Despite the strides made in TL and SSL studies, emerging research suggests that the ubiquitous ImageNet \citep{russakovsky2015imagenet} dataset may not be optimal for pre-training model in medical field \citep{raghu2019transfusion}. The inherent dissimilarity between natural images in ImageNet and medical images raises concerns about domain mismatch adversely affecting downstream task training. A 'medical version ImageNet' is conspicuously absent, compelling researchers to turn to SSL methods that do not need manual labels during pre-training. However, there is still a notable gap in understanding how well these pre-train methods perform in various scenarios.

Observed from our previous study \citep{zhao2023comparison}, the TL methods that operate within a supervised learning framework have yield performance gaps towards different data types, sizes, and domain environments when compared to the SSL methods that are performed under the unsupervised learning setting. Raising the necessity for a nuanced understanding of both methods and reducing time and computational costs in numerous research endeavours. 

Some researchers have begun to take note of this knowledge gap between TL and SSL methods and have endeavoured to address this issue. \cite{yang2020transfer} have conducted a comparative experiment of the two pre-train methods to analyse the impact of domain similarity on the two pre-train methods, across various source and target domains, covering daily objects, natural environments, and even medical data. However, their research did not consider the differences and influences of image colour information, target dataset size, and data imbalance issues, which are important conditions in the training process and have often faced challenges in the medical field. Moreover, they did not consider the robustness of their model, since it is vital for models to provide explainable and reliable answers instead of simply providing the correct answers in the medical field. \cite{azizi2023robust} elucidated the benefits of the SSL method, noting its ability to achieve comparable performance with TL models while using less than 10\% of the target domain data. However, their investigation primarily focused on the development of sophisticated SSL architectures rather than undertaking a comprehensive comparative analysis of the two methodologies.

To tackle this performance gap and enhance the application of both TL and SSL methods to a wider space, this study aims to dive deeper into the factors that influence their performance, and present a new approach to boost these methods in the medical domain. Specifically, a comparative experiment using the TL and SSL methods was designed to identify their effectiveness in different medical imaging modalities. Furthermore, based on the observation from the comparative experiment, we utilised a data-altering method called Deep Convolutional Generative Adversary Network (DCGAN) to solve the data imbalance and scarcity problem and employed a double fine-tuning technique to integrate with the pre-training process to address the domain mismatch issues between source data and target medical data. Furthermore, one Explainable Artificial Intelligence (XAI) technique called Grad-CAM was used to analyse the reliability and robustness of two pre-trained models, providing evidence for the pre-trained models in learning useful features. The knowledge gained from this study will deepen the understanding of the knowledge transfer process, and pave the way for future advancements in both the TL and SSL methods. Thus, we would like to emphasise the contributions of this study to knowledge:
\begin{enumerate}
\item \textbf{Comparative Evaluation}: To the best of our knowledge, this is the first study to compare the strengths and weaknesses of TL and SSL methods across both colourful and grey-scale medical datasets.
\item \textbf{Performance Insights}: Our comparative experiment across different training environments revealed a nuanced performance gap, highlighting the influence of specific medical challenges on these methods.
\item \textbf{Novel Approach}: We proposed a unique solution that integrates data-altering and double fine-tuning techniques with two pre-trained models to address key issues, such as data imbalance, domain mismatch, and data scarcity.
\item \textbf{Improved Results}: The model trained using our proposed approach, along with the two pre-train methodologies, outperformed current state-of-the-art CNN models on various publicly available medical datasets.
\end{enumerate}

 The introduction section provides background and contributions of our research; the rest of the paper is organised as follows: The second section delves into the current applications and research of DL, TL, and SSL methods in the medical field and reveals several common limitations within pre-train methods. The third section then presents the details of the datasets used and outlines the methods proposed in this study, elucidating the experimental process. The fourth section compiles the results of the experiment and performs a comparative analysis of the performance of the two pre-train methods, demonstrating their effectiveness in diverse medical datasets. Finally, the last section summarised the findings of this study, providing a guideline for selecting appropriate pre-train methods in various medical application scenarios. 

\section{Literature Review}

In this section, we briefly review the challenges of deep learning in the field of medical imaging and stress the motivation of this study. Additionally, we highlight recent constraints encountered in the application of transfer learning and self-supervised learning within medical contexts to emphasise the importance of our study.

\subsection{Deep Learning for Medical Imaging Tasks}
Deep learning has emerged as a highly promising technique for various analytical tasks, encompassing classification, segmentation, and generation. Within the medical field, the primary objective of DL models is to detect and locate diseases quickly and efficiently, assisting physicians in rapid diagnoses. Although DL models demonstrate considerable prowess in medical tasks, the performance of DL models in medical imaging tasks is fundamentally dependent on the volume of the dataset, and data scarcity remains a persistent challenge in the medical field \citep{wang2021overview}. The issue of lacking training data comes from various reasons, such as privacy concerns, the utilisation of different medical imaging techniques, and the high cost of acquiring medical images \citep{razzak2018deep}. In addition, the requirement for professional knowledge to annotate medical data has contributed to this scarcity. Consequently, data scarcity in the medical field has limited the development and application of DL technique, requiring more efforts to improve.

\subsection{Transfer Learning for Medical Imaging Tasks}
Transfer learning, also known as knowledge transfer, is a machine learning (ML) technique that utilises the knowledge acquired from one task to improve performance on a related task. The objective of the TL method is to enrich the learnt feature of the pre-trained model and reduce its data dependence \citep{pan2010survey}. With the proliferation of deep learning, transfer learning has become integral to various applications, particularly in the medical field. Numerous studies have explored the factors that contribute to the success of transfer learning in CNNs. These factors include choosing an optimal deep learning architecture for guaranteed model performance \citep{kornblith2019better}, choosing source domains that share a close distribution to the target domain to accelerate better transfer performance \citep{zhang2022survey}, and increasing the size of the source datasets and pre-trained model parameters for richer knowledge transfer \citep{kolesnikov2020big}. In particular, a central drawback of TL is the requirement for pre-training with a large amount of labelled data \citep{huh2016makes}, which requires a substantial pretext dataset such as ImageNet. Nevertheless, despite the positive results reported with pre-trained ImageNet models on colourful medical datasets \citep{morid2021scoping, patil2024automatic}, questions have arisen regarding the suitability of ImageNet for the grey-scale medical datasets \citep{raghu2019transfusion}. Evidence suggests that the transfer learning method does not consistently produce favourable results in the medical domain \citep{alzubaidi2020towards}, as ImageNet representations do not align well with those of medical images, particularly grey-scale ones.

\subsection{Self-supervised Learning for Medical Imaging Tasks}
Self-supervised learning is a recently appeared unsupervised pre-train method that employs model-augmented data as pseudo-labels to provide supervision signals during the pre-training process. Compared to the transfer learning method, self-supervised learning appears particularly attractive in the medical field, offering an effective solution to use large amounts of unlabelled data \citep{shurrab2022self}. In particular, the self-supervised learning model exhibits improved accuracy performance by incorporating more complex auxiliary tasks. Some studies have indicated that self-supervised learning models, particularly those employing contrastive methods, tend to learn invariant features by maximising similarities between raw input data and self-generated pseudo labels \citep{ericsson2022self}. Despite certain self-supervised learning studies that have utilised ImageNet with self-supervised pre-train methods for medical imaging tasks and achieved positive results \citep{truong2021transferable}, there remains a lack of conclusive evidence to ascertain whether pre-trained models with self-supervised learning methods can consistently outperform transfer learning methods in medical imaging tasks and potentially become a new standard in this domain.

\begin{table*}[h!]
\centering
\begin{adjustbox}{width=1\textwidth}
\begin{tabular}{|c c c|}
\hline
\rowcolor{brown!40}\textbf{Reference} &\textbf{Method} & \textbf{Problem description}\\
\hline
\cite{del2023applications} & Self-Supervised Learning & Several factors within SSL training process remain unclear, including \\
& &  choose pretext tasks, data aggregation, and acquiring target information \\
\rowcolor{brown!40}\cite{wu2023self} & Self-Supervised Learning & The high computational resource cost of SSL training process makes it  \\
\rowcolor{brown!40} & &  hard to fully take the place of TL. \\
\cite{zhao2023comparison} & TL \& SSL & Unclear standard of utilising TL and SSL under different scenarios.  \\
\hline
\rowcolor{brown!40}\multicolumn{3}{|c|}{\textbf{Problem Summary: lacking understanding between TL and SSL}}\\
\rowcolor{brown!40}\multicolumn{3}{|c|}{\textbf{may cause confusion in method utilisation} } \\
\hline
\cite{romero2020targeted} & Transfer Learning & Pre-trained TL model using ImageNet dataset does not fit X-ray\\
& &    based classification tasks.\\
\rowcolor{brown!40}\cite{alzubaidi2021novel} & Transfer Learning & Pre-trained TL model using general image datasets does not \\
\rowcolor{brown!40}& & fit with target medical images.\\
\cite{atasever2023comprehensive} & Transfer Learning & Using a light-weight model trained with target dataset directly can \\
& & outperform the pre-trained TL model using natural images.\\
\rowcolor{brown!40}\cite{tan2024self} & Self-Supervised Learning & Pre-trained SSL model using natural images does not perform well with\\
\rowcolor{brown!40}& &  target COVID-19 samples and need further guidance from user.\\
\hline
\multicolumn{3}{|c|}{\textbf{Problem Summary: domain discrepancy during pre-training will  }}\\
\multicolumn{3}{|c|}{\textbf{degrade pre-trained model's performance }}\\
\hline
\rowcolor{brown!40}\hline\cite{muljo2023handling} & Transfer Learning & Utilising pre-trained TL model does not bring significant improvement to \\
\rowcolor{brown!40}& &  the target medical dataset with an imbalanced sample distribution. \\
\cite{zhang2023dive} & Self-Supervised Learning & The imbalanced source and target datasets lead to poor model \\
& &   performance even after self-supervised pre-training.\\
\hline
\rowcolor{brown!40}\multicolumn{3}{|c|}{\textbf{Problem Summary: neither TL or SSL methods show improved}}\\
\rowcolor{brown!40}\multicolumn{3}{|c|}{\textbf{  performance towards imbalanced datasets}}\\
\hline
\cite{qayyum2020secure} & Transfer Learning &  The complexity of model and pre-training process makes it hard to\\
& &  understand the results and reduce the reliability of predictions.   \\
\rowcolor{brown!40}\cite{schiappa2023self} & Self-Supervised Learning & The pre-training process of SSL model is fully unsupervised, which \\
\rowcolor{brown!40}& & raised the concern for whether the model have fully understand the  \\
\rowcolor{brown!40}& &  target dataset or is making predictions based on random factors.\\

\hline
\multicolumn{3}{|c|}{\textbf{Problem Summary: the complexity of knowledge transferring }}\\
\multicolumn{3}{|c|}{\textbf{ process raised concerns of model reliability}}\\
\hline
\end{tabular}
\end{adjustbox}
\caption{Four main issues that constrained the application of pre-train methods in the medical field are summarised here: 1. the performance gap between TL and SSL in different data modalities, 2. the domain mismatch gap between source and target domain, 3. the challenge of data imbalance scenarios, 4. the difficulty in model explainability and analysis. }
\label{Limitations of recent pretrained methods}
\end{table*}

\subsection{Recent Limitations of Transfer Learning and Self-supervised Learning for Medical Imaging Tasks}

Although both TL and SSL methods have been applied in the medical domain and were reported with positive results, several challenges impede their advancement. To better identify the challenges, we conclude some common issues related to these two methods (see table~\ref{Limitations of recent pretrained methods}).   

In summary, we want to stress four main problems that occurred frequently during the pre-training process, including the lack of understanding between TL and SSL methods, the performance gap caused by domain discrepancy and imbalanced data, and the reliability concern of the pre-trained models. The first problem occurs because the TL and SSL methods are both rooted in DL architectures, and there has been a longstanding tendency to perceive them as interchangeable and overlook their distinct performance across various types and sizes of data. The second problem is domain discrepancy and unbalanced data, common data issues within the medical domain. However, recent studies revealed that depending on a single pre-train method can hardly handle the data issue, leaving a novel research gap. The last problem comes from the black box structure of DL architectures, which compounds the challenge of understanding model predictions. Also, the knowledge-transferring process of the pre-training model has increased the difficulty of understanding the influence of source data, making the model's prediction particularly hard to explain.

All of the aforementioned problems related to TL and SSL methods have reduced their reliability and raised concerns about whether AI technology should be widely used in real-world medical applications. Addressing these problems has significant value in boosting AI applications in the medical field and providing patients with faster and more accurate diagnoses.

\section{Comparative Experiment and Evaluation}

\begin{figure*}[htbp]
\centerline{\includegraphics[width=1\textwidth]{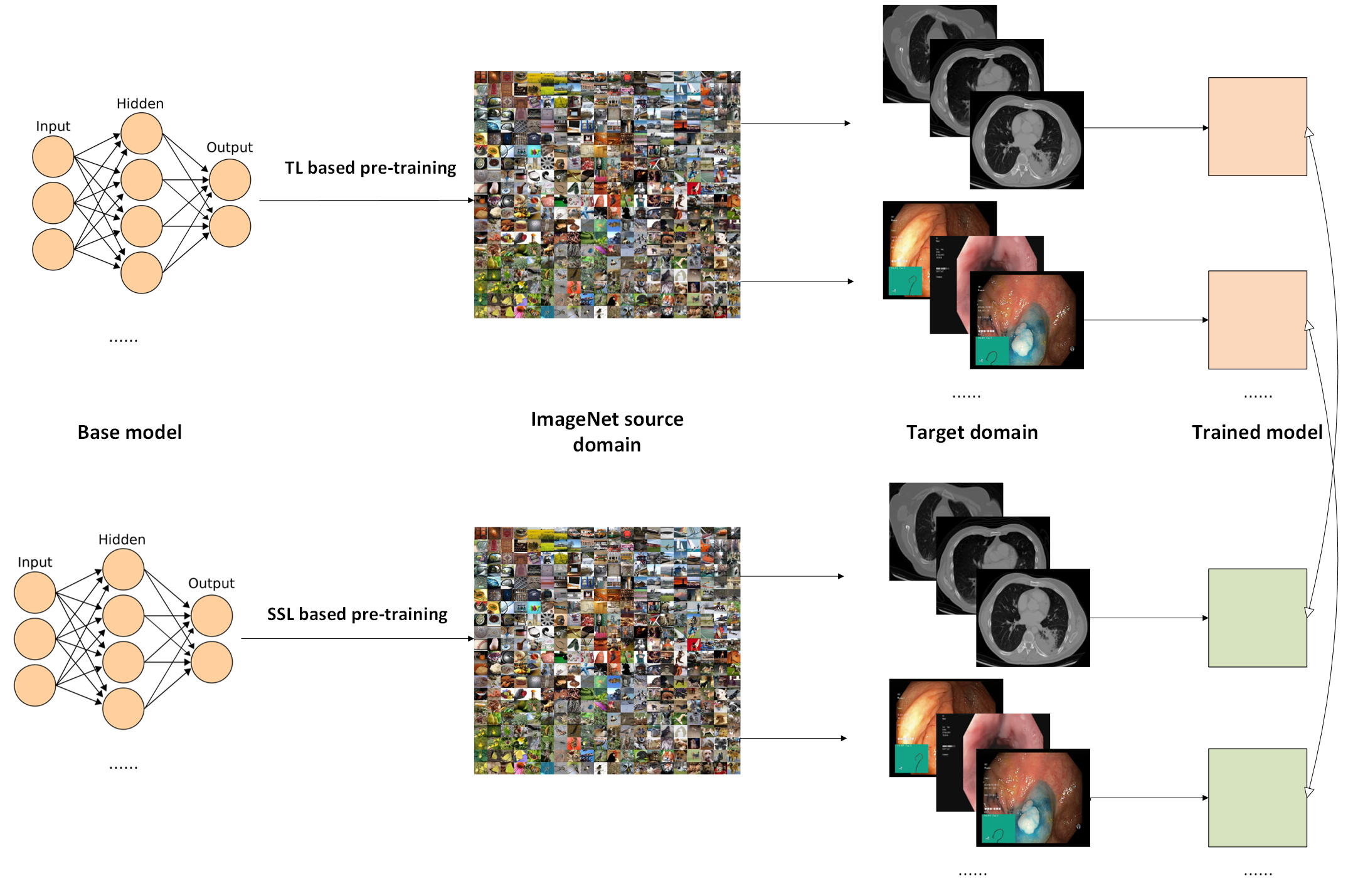 }}
\caption{The base models are pre-trained with TL and SSL methods independently and fine-tuned to the same target datasets, the test results of trained models are then listed and compared in the following table.}
\label{Comparison experiment workflow}
\end{figure*}

In this section, we conduct a comparative experiment between TL and SSL methods to examine their performance differences within the medical field, highlighting key influencing factors. Four widely used CNN models, ResNet50, InceptionResNetV2, Xception, and MobileNetV3, were selected as the base architectures. Additionally, four medical imaging datasets, including both grey-scale and colour images from various imaging modalities, were used as the target domain datasets. To reduce the influence of irrelevant factors, all the base models are pre-trained independently with TL or SSL methods using ImageNet as a source domain and fine-tuned to the target medical dataset. The overall experiment process is shown in Fig.~\ref{Comparison experiment workflow}

\subsection{Pre-train Methods}

We follow a supervised paradigm for the TL setting to pre-train our base model. First, the model is pre-trained on the ImageNet dataset as the source domain, and then these pre-trained parameters are transferred as an initial setup to the target medical datasets. During fine-tuning with the target dataset, we freeze 60\% of the model's bottom layers to retain the foundational knowledge from ImageNet, while updating the remaining top layers to adapt to and learn from the target data. All trained models are evaluated using a separate subset of the target datasets as testing samples.

\begin{figure}[htbp]
\centerline{\includegraphics[width=0.9\textwidth]{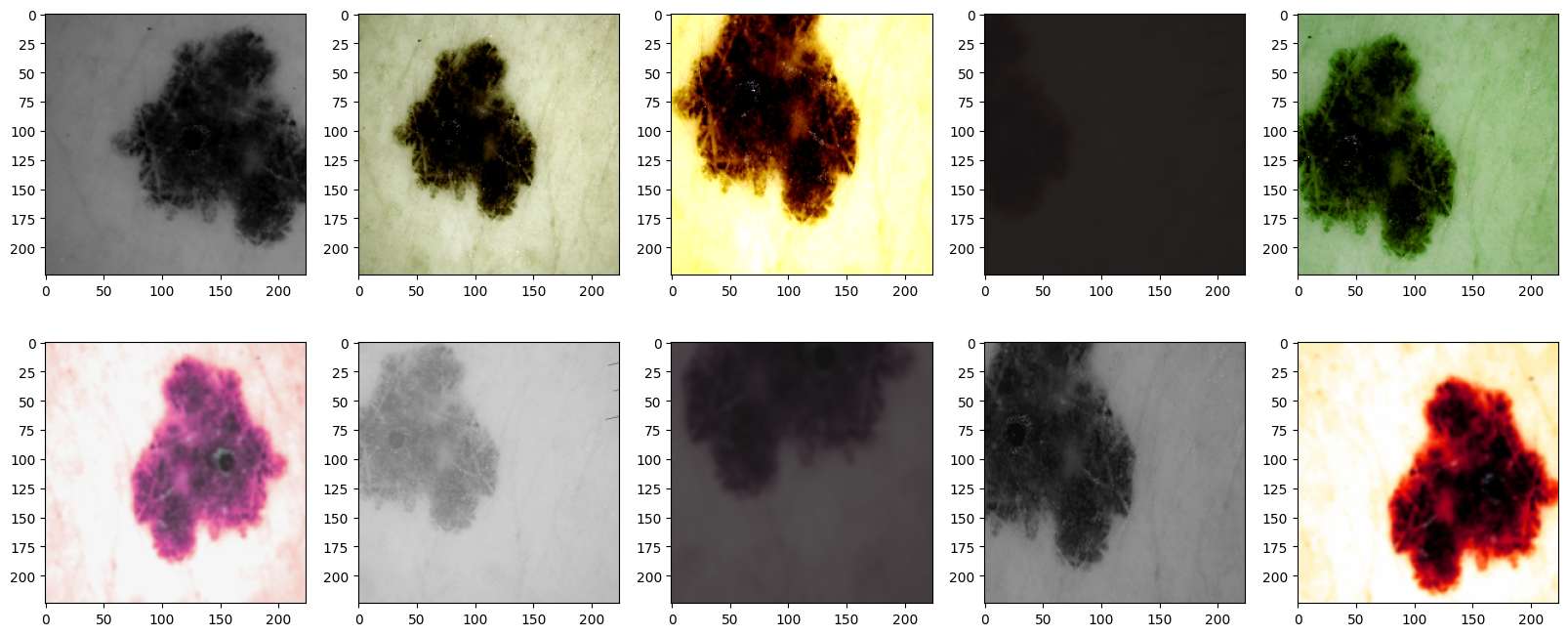 }}
\caption{Self-generated pseudo label sample.}
\label{Pseudo label sample}
\end{figure}

For the SSL setting, we use the Simple Framework for Contrastive Learning of Visual Representations (SimCLR) method to pre-train our base model \citep{chen2020simple}. The base models are trained with the same source data, but under the SimCLR paradigm. In this setup, the model generates image pairs through data augmentation techniques such as resizing, flipping, zooming, rotating, and adjusting brightness and hue which serve as pseudo labels for training and support contrastive loss calculation. Fig.~\ref{Pseudo label sample} shows an example of a pseudo-labeled image generated in this process.

The SimCLR method further utilised a different loss function called NT-Xent to calculate the training loss by contrasting the representations of positive pairs (similar images) and concurrently creating negative pairs (dissimilar images). The equation of the loss function is as follows. 

\begin{small}
$$ l_{i,j} = -\log\frac{exp(sim(z_i,z_j)/\tau)}{\sum^{2N}_{k=1}1[k\neq i]exp(sim(z_i,z_k)/\tau) } $$
\end{small}

The positive pair of example (i,j) uses $1[k\neq i]$ as an indicator function to evaluate if $k\neq i$, and $\tau$ denotes a temperature parameter while the final loss is computed by summing all positive pairs and divide by 2 x N = views x batch-size. The loss function encourages the model to bring positive pairs closer within the embedding space, while pushing negative pairs further apart, thereby enhancing the learning from input representations. Following the SimcLR method, the base models are pre-trained with self-supervised signals. The fine-tuning steps of SSL-based models are similar to these TL-based models, using 40\% top layers to train with target medical datasets using a supervised paradigm to acquire target information.

\subsection{Base Architectures}

The objective of this experiment is to compare the effectiveness of TL and SSL methods in the medical field and provide evidence for their strengths and weaknesses. To reduce the influence of base model performance, we selected different base models and trained them with the same pre-train method and target datasets. In this subsection, we briefly introduce the selected base models and explain how they work. 

1. The ResNet50 architecture \citep{he2016deep} is a well-performed CNN model and has been widely tested and applied to different domains. The structure of ResNet consists of 48 convolutional layers, one max pool layer, and one average pool layer. Each convolutional layer uses a bottleneck design for the building block which consists of two 1x1 convolutions as the bottleneck to process the input and output features, reducing the number of parameters and matrix multiplications. The identity mapping used in ResNet allows the model to bypass a CNN weight layer if the current layer is not necessary, which also helps avoid overfitting problems to the training set. 

2. The InceptionResNetV2  \citep{szegedy2017inception} is a mixed model that utilises residual layers to reinforce the inception block and achieve better performance. The structure of the residual inception block contains two pipelines from residual layers and the inception block, while the input data are processed parallel and concatenated later to improve the model's efficiency. The InceptionResNetV2 model provides improved performance compared to the original Inception model and reduces computational cost, making it easier to implement. 

3. The Xception model \citep{chollet2017xception} is a widely used DL architecture known for its high performance. The Xception module is constructed as a series of three flow structures, each incorporating batch normalisation (BN), rectified linear units (ReLU) activation functions, and separable convolutional kernels in depth. Within the Xception architecture, its 36 convolutional layers are organised into 14 modules. Each module operates through sequential processing: first through the entry flow, then via the middle flow, and ultimately undergoing further processing in the exit flow. 

4. The MobileNetV3 model \citep{howard2019searching} is a deep neural network architecture designed with efficient depthwise separable convolutions as a streamlined alternative to traditional convolution layers. Depthwise separable convolutions consist of two parts: a lightweight depthwise convolution layer for spatial filtering and a pointwise convolution layer for feature generation. This architecture enables MobileNetV3 to support a more efficient training process and offers a flexible implementation environment.

\subsection{Datasets}

This part overviews the target datasets (see Table \ref{Dataset detail}). Considering that grey-scale and colourful images are both commonly used in the medical field, we want to select multiple datasets, including grey-scale samples and colourful samples, to test the influence of colour information on the pre-trained models. Furthermore, we consider datasets with limited samples and imbalanced data distribution to simulate real-world scenarios and test both pre-train methods' stability towards them. Specifically, four representative data sets were chosen as target data sets in this evaluation experiment, including two greyscale data sets named BusI and Lung Cancer CT and two colourful data sets Kvasirv2 and EyePacs. More details of each dataset are listed below.

\begin{table*}[h!]
\centering
\begin{adjustbox}{width=0.9\textwidth}
\begin{tabular}{|c|c|c|c|c|}
\hline
\textbf{Dataset} &\textbf{Image Count} & \textbf{Image Modality} & \textbf{Imbalanced Data} & \textbf{Classes}\\
\hline
\multicolumn{5}{|c|}{\textbf{Grey-scale Image Datasets}}\\
\hline
BusI  & 780 & Ultrasound & Yes & 3\\
Chest CT  & 900 & CT & Yes &4\\
\hline
\multicolumn{5}{|c|}{\textbf{Colourful Image Datasets}}\\
\hline
Kvasirv2  & 8000 & Gastrointestinal tract & No&8\\
EyePacs & 6000 & Fundus & No &2\\
\hline
\end{tabular}
\end{adjustbox}
\caption{\textbf{Summary of target datasets}: The BusI and Lung Cancer datasets that are based on grey-scaled images also have data imbalance issues and limited size. The Kavasirv2 and EyePacs datasets are based on colourful images with bigger sizes and balanced sample distribution.}
\label{Dataset detail}
\end{table*}

1. BusI \citep{al2020dataset} is a breast ultrasound image dataset for grey-scale breast cancer detection images. The dataset is an imbalanced dataset consisting of 780 images that are separated into three categories: Benign, Malignant, and Normal. The benign class comprises 437 images, 210 images belong to the malignant class, and the remaining 133 samples are normal images. This dataset faces three main challenges: limited colour channels, limited sample size, and imbalanced sample distribution. 

2. Chest CT Dataset \citep{ha2020chest} is a lung cancer dataset collected from the Kaggle website, comprising 900 grey-scaled CT images belonging to four classes, representing three chest cancer types and a normal type: adenocarcinoma, large cell carcinoma, squamous cell carcinoma, and normal cell. The adenocarcinoma class possess 326 image samples, the large cell carcinoma class holds 163 image samples, the normal class has 159 image samples, and the squamous cell carcinoma class reserved 252 image samples. Similar to the BusI dataset, it is also constrained by its limited colour space and imbalanced data distribution.

3. KvasirV2 dataset \citep{pogorelov2017kvasir} contains 8000 samples from the gastrointestinal (GI) tract. The dataset has 8 classes related to anatomical landmarks and endoscopic polyp removal, listed as dyed lifted polyps, dyed resection margins, esophagitis, normal cecum, normal pylorus, normal Z line, polyps, and ulcerative colitis. This dataset is balanced as each class consists of 1000 sample images and all of the images have been annotated and sorted by experienced endoscopists. 

4. EyePacs dataset \citep{kiefer2023automated} is a generic glaucoma dataset composed of 6000 fundus images, divided into two classes of referable glaucoma (RG) and non-referable glaucoma (NRG). The data distribution of each class is balanced, and data samples are based on colourful images. This dataset only contains two classes and can serve as a binary classification task, while the other three datasets are based on multi-class classification tasks.

\subsection{Results and Analysis}

The training preparation stage includes downloading pre-trained ImageNet weights, preprocessing the input data, and setting specific hyperparameters for the base models. To minimise differences caused by varying hyperparameters or input data, all base models use the same hyperparameter setup and process the original images from the target dataset by resizing them to 224x224 pixels. Specifically, we used the Adam optimizer with a learning rate of 0.001, a weight decay rate of 0.001, and a batch size of 64 for 50 epochs. The loss function is categorical cross-entropy, with a learning rate annealer that decays by 0.1 each time and a minimum learning rate of 0.00001 if test loss fails to decrease after three epochs.

\begin{table*}[h!]
\centering
\begin{adjustbox}{width=1\textwidth}
\begin{tabular}{|c |c c c c c c |}
\hline
\textbf{Dataset} & \textbf{Base Model} &\textbf{Method} &\textbf{Acc(\%)}  & \textbf{Sen(\%)} & \textbf{Pre(\%)} & \textbf{F1(\%)}\\
\hline
\multirow{8}{4em}{BusI}  &  ResNet & TL & 69.20  & 69.16 & 79.21 & 69.28 \\
& InceptionResNet & TL & 68.34  & 68.15 & 68.67 & 68.82\\
 & Xception & TL& 69.20  & 69.16 & 69.21 & 69.28 \\
  & MobileNet & TL& 67.40  & 67.40 & 67.42 & 67.40 \\ 
  \hhline{|~|------|} 
& ResNet & SSL & 74.16  & 74.16 & 79.58 & 73.71 \\
 & InceptionResNet& SSL& 78.35  & 78.35 & 78.35 & 78.35\\
 & Xception & SSL& 81.22  & 80.76 & 81.56 & 80.87 \\
  & MobileNet & SSL& 80.96  & 80.96 & 85.24 & 81.44 \\
\hline
\multirow{8}{4em}{Chest CT}  &ResNet & TL &85.23  & 86.60 & 85.40 & 85.99\\
 & InceptionResNet& TL& 87.02  & 87.02 & 87.02 & 87.01\\
 & Xception & TL& 87.30  & 86.60 & 90.20 & 88.11 \\
  & MobileNet & TL& 86.48  & 86.60 & 86.45 & 86.52 \\
  \hhline{|~|------|} 
& ResNet & SSL & 93.02  & 94.18 & 93.03 & 93.42 \\
 & InceptionResNet& SSL& 92.88  & 93.32 & 93.03 & 93.17\\
 & Xception & SSL& 94.41  & 94.18 & 93.03 & 93.42 \\
  & MobileNet & SSL& 92.40  &92.40 & 92.49 &92.23\\
 
\hline
\multirow{8}{4em}{Kvasirv2}  &ResNet & TL&90.23  & 90.90 & 91.00 & 90.87  \\
 & InceptionResNet& TL& 91.22  &91.24 & 91.20 &91.21\\
 & Xception & TL& 91.50  & 90.90 & 91.00 & 90.87 \\
  & MobileNet & TL& 91.62  &91.62 & 92.22&91.91 \\
  \hhline{|~|------|} 
& ResNet & SSL & 82.93  & 82.93 & 83.03 & 82.89 \\
 & InceptionResNet& SSL& 84.32  &84.36 & 84.66 &84.52\\
 & Xception & SSL& 85.81  &85.81 & 82.96&84.36 \\
  & MobileNet & SSL& 78.30  &82.49 & 82.66 &77.51 \\
\hline
\multirow{8}{4em}{EyePacs} & ResNet & TL&88.57  & 88.60 & 89.21 & 88.90\\
 & InceptionResNet& TL&92.10  & 92.10 & 92.00 & 92.00\\
 & Xception & TL& 91.20  &91.10 & 91.10&91.00 \\
  & MobileNet & TL& 90.76  &90.76 & 90.76&90.76 \\
  \hhline{|~|------|} 
& ResNet & SSL & 80.46  & 80.47& 80.46 &80.46 \\
 & InceptionResNet& SSL& 84.56  &84.49 & 82.87 &83.85\\
 & Xception & SSL& 83.20  &83.20& 83.20&83.20 \\
  & MobileNet & SSL& 83.23  &83.44 & 83.46&83.44 \\
\hline
\end{tabular}
\end{adjustbox}
\caption{Each target dataset is split by 8:2 to serve for fine-tuning and testing use. The testing results of each base model group are based on TL and SSL trained models using the test sets.}
\label{Comparative experiment}
\end{table*}

After preparing the comparative experiment and training base models on each target dataset, we tested and compared their performance using several evaluation metrics. The results are shown in Fig.\ref{Comparative experiment}, while the evaluation metrics we used to analyse them included Accuracy, Recall, Precision, and F1-score. The calculation equations of these performance measurements are shown as follows. TP means true positive, TN is true negative, and they represent the number of correctly classified negative and positive instances, respectively. On the other hand, FP is false positive, and FN stands for false negative, which denotes the number of misclassified positive and negative cases, respectively.

\begin{small}
$$ Accuracy = \frac{TN+TP}{TN+FN+TP+FP} $$
\end{small}

\begin{small}
$$ CrossEntropy-Loss = -\sum_{c=1}^My_{o,c}\log(p_{o,c}) $$
\end{small}

\begin{small}
$$ Sensitivity = Recall = \frac{TP}{TP+FN} $$
\end{small}

\begin{small}
$$ Precision = \frac{TP}{TP+FP} $$
\end{small}

\begin{small}
$$ F1 = 2 * \frac{Precision * Recall }{Precision + Recall} $$
\end{small}

Based on the results of the comparative experiment, despite the performance difference caused by the architecture themselves, the pre-trained models following the TL showed stronger performance on the Kvasirv2 and EyePacs datasets, with an average accuracy 8.3\% higher than that of the models using SSL method. However, this trend reversed on the BusI and Chest datasets, where SSL-based models achieved an average accuracy that was 10.3\% higher than the TL-based models. 

As mentioned previously, a key difference between the BusI-Chest CT dataset group and the Kvasirv2-EyePacs dataset group is in the colour channels, as the samples from BusI-Chest CT are based on greyscale images, and Kavasirv2-EyePacs samples are based on colourful images. Furthermore, the BusI-Chest CT group presents more challenges, such as data imbalance and limited data availability. From these observations, we infer that differences in colour channels may contribute to the performance gap between TL- and SSL-based models, while data imbalance and data scarcity could also be significant factors. To further explore their impact, we extend our comparative experiment by applying data augmentation and downsampling techniques to simulate these conditions and observe the resulting changes in the performance of each pre-trained model.

\subsection{Extended Test Results}

Specifically, we aim to extend our comparative experiment in two directions: (1) slightly augment and rebalance the target datasets to simulate a scenario without data imbalance, and (2) create a smaller target dataset to examine the impact of limited dataset size on pre-train methods.

For the first direction, we applied data augmentation techniques such as rotation, flipping, and zooming in the target datasets to increase the sample sizes of imbalanced classes. The same augmentation was also applied to balanced datasets for comparison. For the second direction, we used down-sampling by randomly reducing the sample count in each class. Details on the sample size adjustments for each dataset are provided in Tables.~\ref{Augmentated datasets size} and \ref{Downsampled datasets size}.

This time, we selected the Xception model as the base model, given its superior performance compared to other models with both TL and SSL methods. By comparing the performance of this model on both augmented and down-sampled datasets, we can isolate the effects of data imbalance and colour information, gaining a clearer understanding of how data limitations affect pre-trained models. Furthermore, by comparing these results with those from the previous experiment, which did not use any data-altering techniques, we can derive additional insights into the impact of colour information and data imbalance on model performance.

\begin{table}[h!]
\centering
\begin{adjustbox}{width=0.95\textwidth}
\begin{tabular}{c c c c}
\hline
\textbf{Dataset} & \textbf{Method} & \textbf{Category} & \textbf{Sample Number} \\
\hline
\multirow{3}{4em}{BusI}& Augmentation & Train Set& 397 images each class\\
& - & Test Set& 40 images each class\\
\cmidrule(lr){4-4}
& & & Total: 1311 images \\
\hline
\multirow{3}{4em}{ChestCT}& Augmentation & Train Set& 236 images each class\\
 & - & Test Set& 45 images each class\\
\cmidrule(lr){4-4}
& & & Total: 1124 images \\
\hline
\multirow{3}{4em}{Kvasirv2} & Augmentation & Train Set& 1600 images each class\\
& - & Test Set& 200 images each class\\
\cmidrule(lr){4-4}
& & & Total: 14400 images \\
\hline
\multirow{3}{4em}{EyePacs}  & Augmentation & Train Set& 5000 images each class\\
 & - & Test Set& 500 images each class\\
\cmidrule(lr){4-4}
& & & Total: 11000 images \\
\hline
\end{tabular}
\end{adjustbox}
\caption{Augmentated target datasets aligned the sample number of each class to its largest class, increasing the size of that target dataset and reducing the influence of data imbalance at the same time.}
\label{Augmentated datasets size}
\end{table}

\begin{table}[h!]
\centering
\begin{adjustbox}{width=0.95\textwidth}
\begin{tabular}{c c c c}
\hline
\textbf{Dataset} & \textbf{Methods} & \textbf{Category} & \textbf{Sample Number} \\
\hline
\multirow{3}{4em}{BusI}& Down-sample & Train Set& 108 images each class\\
& - & Test Set& 25 images each class\\
\cmidrule(lr){4-4}
& & & Total: 399 images \\
\hline
\multirow{3}{4em}{ChestCT} & Down-sample & Train Set& 127 images each class\\
 & - & Test Set& 32 images each class\\
\cmidrule(lr){4-4}
& & & Total: 636 images \\
\hline
\multirow{3}{4em}{Kvasirv2}& Down-sample & Train Set& 100 images each class\\
 & - & Test Set& 25 images each class\\
\cmidrule(lr){4-4}
& & & Total: 1000 images \\
\hline
\multirow{3}{4em}{EyePacs} & Down-sample & Train Set& 500 images each class\\
 & - & Test Set& 100 images each class\\
\cmidrule(lr){4-4}
& & & Total: 1200 images \\
\hline
\end{tabular}
\end{adjustbox}
\caption{Down-sampled target datasets aligned the sample number of each class to its smallest class, simulating the severe environment when only limited data samples are avaliable.}
\label{Downsampled datasets size}
\end{table}

\begin{table*}
\centering
\begin{adjustbox}{width=1\textwidth}
\begin{tabular}{|c |c c c c c c |}
\hline
\textbf{Dataset} & \textbf{Data-altering Method} &\textbf{Method} &\textbf{Acc(\%)}  & \textbf{Sen(\%)} & \textbf{Pre(\%)} & \textbf{F1(\%)}\\
\hline
\multirow{4}{4em}{BusI}  & Augmentation & TL& 78.30  & 82.49 & 82.66 & 77.51 \\
  & Down-sampling & TL& 64.00  & 64.00 & 69.17 & 64.19 \\ 
  \hhline{|~|------|} 
 & Augmentation & SSL& 83.10  & 79.40 & 83.21 & 81.05\\
 & Down-sampling & SSL& 72.00  & 72.00 & 73.48 & 71.88 \\

\hline
\multirow{4}{4em}{Chest CT}  & Augmentation & TL& 93.30  & 93.32 & 93.69 & 93.39 \\
  & Down-sampling  & TL& 62.50  & 68.28 & 68.14 & 67.92 \\
  \hhline{|~|------|} 
 & Augmentation & SSL& 95.27  & 95.56 & 95.67 & 95.59 \\
  & Down-sampling & SSL& 91.40  &91.40 & 91.49 &91.23\\
 
\hline
\multirow{4}{4em}{Kvasirv2} & Augmentation & TL& 93.40  & 93.37 & 93.37 & 93.35 \\
  & Down-sampling & TL& 70.50  &70.50 & 71.28&70.65 \\
  \hhline{|~|------|} 
 &  Augmentation & SSL& 87.22  &87.34 & 87.24&87.27 \\
  & Down-sampling & SSL& 78.00  &78.00 & 77.80 &77.76 \\
\hline
\multirow{4}{4em}{EyePacs} &Augmentation & TL& 92.74  &92.74 & 92.74&92.73 \\
  &Down-sampling & TL& 88.00  &88.00 & 88.38 & 87.90 \\
  \hhline{|~|------|} 
 & Augmentation & SSL& 85.65  &85.66& 85.67&85.66 \\
  & Down-sampling & SSL&75.50  &75.50 & 75.50&75.50 \\
\hline
\end{tabular}
\end{adjustbox}
\caption {The Xception model pre-trained separately with TL and SSL methods is tested on both the augmented and down-sampled target datasets. Results indicate that TL-based models are more sensitive to changes in dataset size. In contrast, SSL-based models demonstrate more stable performance towards challenging environments such as extremely limited datasets. }
\label{Extended comparative experiment}
\end{table*}

The extended comparative experiment results indicate that increasing the target datasets' size can enhance pre-trained models' performance. However, the extent of improvement varies between TL- and SSL-based models: TL-based models showed accuracy gains of up to 11.18\%, while SSL-based models improved by a maximum of 2.45\%. A similar pattern appeared in tests with down-sampled target datasets: TL-based model performance declined sharply, with drops ranging from 5.2\% to 15.3\% as data size decreased. In contrast, SSL-based models experienced a more gradual decline, with reductions between 3\% and 8\%. These findings highlight that data scarcity and data imbalance issues continue to challenge pre-train methods, especially when target datasets are quite small. Notably, these issues are particularly pronounced for TL-based models, underscoring the need for more focus on utilising this method.  

We also compared the performance of the augmented target data sets with the original data sets. Results show that even with a larger, balanced target dataset, the performance gap between TL and SSL based models remains. This finding supports our assumption that colour information influences the performance of pre-trained models. Furthermore, based on observed performance evaluations, TL-based models tend to perform better on colourful image datasets, while SSL-based models excel with grey-scale datasets. The use of ImageNet, a colour-rich dataset, as the source domain, may also explain why both TL and SSL-based models generally perform better on colour-target datasets compared to grey-scale ones.

Through this comparative experiment, we gained deeper insights into TL and SSL methods and highlighted the impact of colour information, data scarcity, and data imbalance on their performance. While current pre-train methods aim to address data scarcity, they prove effective only in specific contexts in the medical field. Moreover, the assumption that SSL methods can fully replace TL methods is unreliable. Although SSL methods show greater stability in more challenging scenarios, TL methods can still outperform SSL when handling colour image samples.

\section{Proposed Approach and Evaluation}

To address the abovementioned limitations in the pre-trained learning process, we propose a novel approach by combining double fine-tuning techniques and data generative model with pre-train methods. In this section, we briefly introduce the workflow of our proposed approach, including its design and training process, and provide an evaluation of the approach based on experiment results and explainable analysis using Grad-CAM technology.  

\begin{figure*}[htbp]
\centerline{\includegraphics[width=1\textwidth, height=3.5cm]{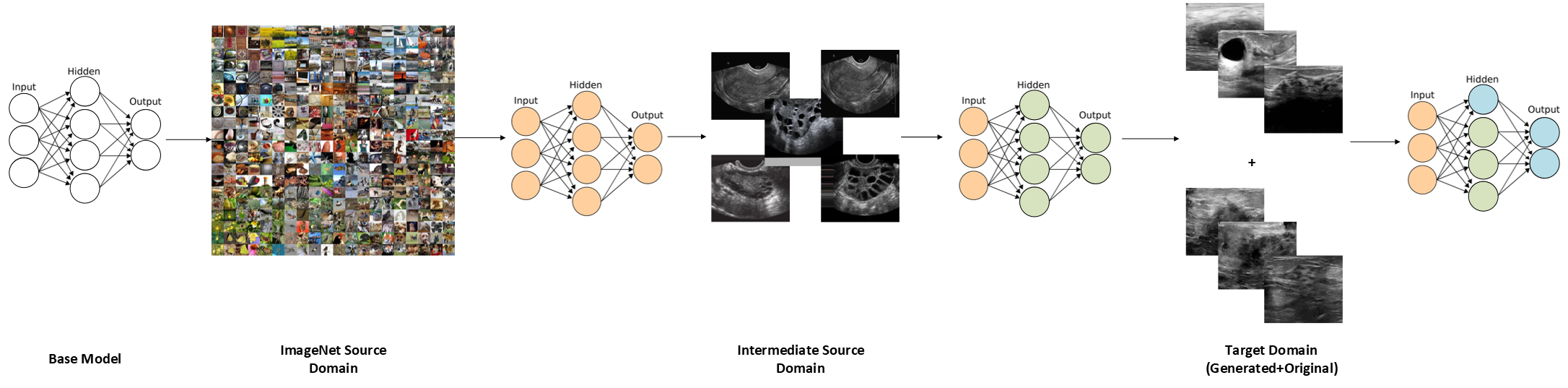 }}
\caption{The base models are pre-trained with TL and SSL methods independently and fine-tuned to the same target datasets, the test results of trained models are then listed and compared in the following table.}
\label{Approach workflow}
\end{figure*}

\subsection{Proposed Methodology}

From our previous experiments, we observed that differences in feature representations, along with image colour, can contribute to domain mismatch between the source and target domains. However, finding a large-scale medical dataset for pre-training is challenging due to the lack of labelled data. Inspired by the works of \cite{tan2017distant} and \cite{niu2021distant}, we aim to address this issue by applying a double fine-tuning technique. This technique bridges the source and target domains by first fine-tuning the pre-trained foundational model on a large training dataset, followed by further fine-tuning on the target domain/task with limited training data. Double fine-tuning allows the base model to use a small, related dataset as an intermediate source, giving it an opportunity to update a subset of parameters and better adapt to the target domain. By retaining knowledge from both the source and intermediate source domains, the base model gains improved generalisability and a reduced risk of data discrepancy.

In our case, we found that using the colour-rich ImageNet as the source domain resulted in unsatisfactory performance on target datasets. Therefore, we plan to double fine-tune the pre-trained model using a medical dataset that shares the same colour characteristics as the target dataset; for example, we’ll use a grey-scale chest X-ray dataset as the intermediate source for a Chest CT target dataset. Additionally, we chose datasets with similar feature representations or from the same medical imaging techniques to reduce domain mismatch risks and enhance performance. The target datasets tested in our previous comparative experiment will be reused here for consistent performance comparison and efficient evaluation.

On the other hand, fine-tuning training is also essential for updating pre-trained models to adapt to target datasets. However, our comparative experiments revealed that this process is often hindered by data imbalance and scarcity issues prevalent in the medical field, leading to suboptimal results. Our findings also confirmed that basic data augmentation techniques, such as flipping and rotation, effectively mitigate data imbalance and scarcity. However, these techniques risk duplication of training data, as augmented samples retain feature representations and distributions identical to the originals. This duplication can bias the model's attention towards frequently appearing features, causing it to overlook rare patterns, leading to misinterpretation of target data and biased predictions.

To mitigate this issue and generate high-quality augmented data, we implemented the DCGAN model\citep{radford2015unsupervised} to create new samples and support the fine-tuning process. DCGAN is an unsupervised generative model capable of learning from target datasets and producing new samples that closely resemble the originals. It consists of two components: a generator, which creates synthetic samples, and a discriminator, which learns to distinguish between real and generated samples. The generator and discriminator engage in a competitive process, enhancing the generator's ability to produce realistic new samples over time. Unlike basic augmentation methods, DCGAN generates images that do not share identical feature distributions with the originals, minimising duplication risks and offering a more robust solution than traditional augmentation techniques.

\begin{figure}[h!]
\begin{subfigure}{0.5\textwidth}
\includegraphics[width=0.6\linewidth, height=3cm]{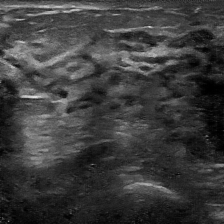} 
\label{normal_13}
\end{subfigure}
\begin{subfigure}{0.5\textwidth}
\includegraphics[width=0.6\linewidth, height=3cm]{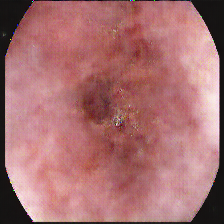}
\label{esophagitis_13}
\end{subfigure}
\begin{subfigure}{0.5\textwidth}
\includegraphics[width=0.6\linewidth, height=3cm]{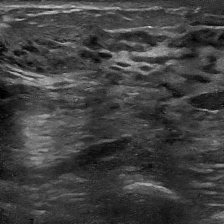}
\label{normal_46}
\end{subfigure}
\begin{subfigure}{0.5\textwidth}
\includegraphics[width=0.6\linewidth, height=3cm]{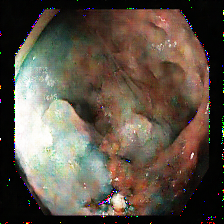}

\label{dyed-lifted_2}
\end{subfigure}
\begin{subfigure}{0.5\textwidth}
\includegraphics[width=0.6\linewidth, height=3cm]{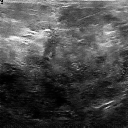}

\label{malignant_13}
\end{subfigure}
\begin{subfigure}{0.5\textwidth}
\includegraphics[width=0.6\linewidth, height=3cm]{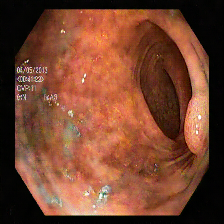}

\label{polyps_10}
\end{subfigure}
\caption{Image samples generated from DCGAN model.}
\label{Gan augmented samples}
\end{figure}

\begin{figure}[h!]
\centerline{\includegraphics[width=1\textwidth]{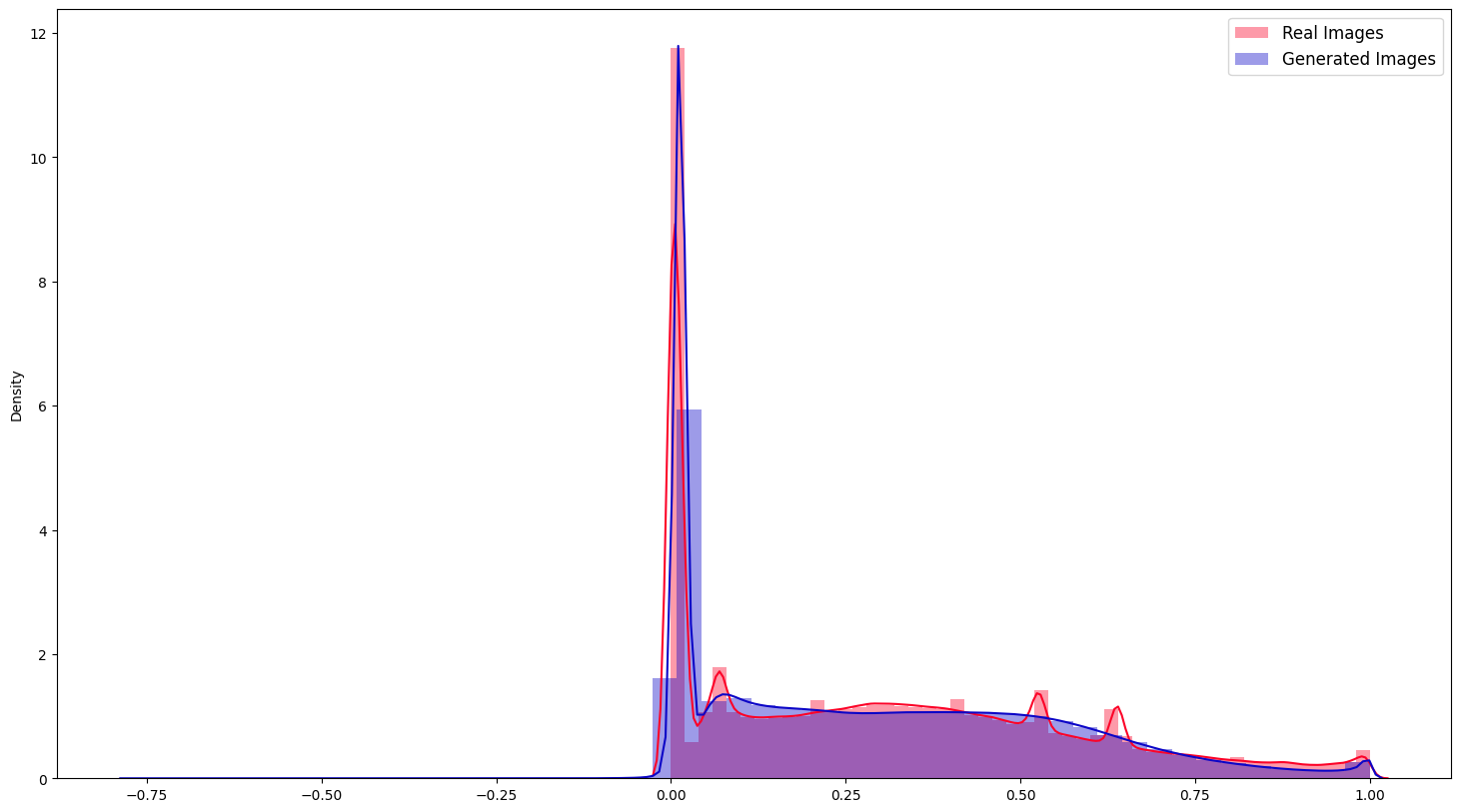 }}
\caption{The feature distribution of generated images vs original samples. The red bar represents original samples, while the blue represents generated images.}
\label{Feature distribution}
\end{figure}

Fig.~\ref{Gan augmented samples} presents examples of model-generated images from the BusI and Kvasirv2 datasets, while Fig.~\ref{Feature distribution} illustrates the feature distribution of these generated images. From these samples and the visualised feature distribution, we observed that although the generated images are physically similar to the target samples, they display a distinct feature distribution. Frequently occurring feature representations in the original samples are less prominent in the generated images, while rarer features maintain their original trends.

The proposed approach combines double fine-tuning and data generation techniques with pre-trained models to address domain mismatch, data imbalance, and data scarcity issues within the medical field. As illustrated in Fig.~\ref{Approach workflow}, the base model is initially pre-trained on ImageNet, then fine-tuned on an intermediate dataset with 30\% of layers frozen, and finally fine-tuned on a mixed target dataset of both generated and original images, with 60\% of layers frozen.

\subsection{Intermediate Source Dataset}

This part introduces the selected intermediate source datasets that match each target dataset. To align with the target datasets' colour characteristics, the chosen intermediate datasets are grouped by the colour channels they poses.

1. Polycystic Ovary Syndrome (PCOS) \citep{chou2021ultra} includes 3,856 grey-scale ultrasound images of ovaries, categorised into infected and non-infected classes. Although the disease symptoms differ from those in the BusI dataset, both datasets use ultrasound imaging, allowing them to share similar feature representations.

2. CT Kidney \citep{islam2022vision} is a comprehensive CT dataset designed to facilitate automatic kidney tumour diagnosis. It contains a total of 12,446 CT whole abdomen and urinary images, focusing on three primary renal disease categories: kidney stones, cysts, and tumours. This dataset was selected because, like the Chest CT dataset, it uses CT imaging, making their samples compatible.

For colourful image training, two medical datasets were chosen to support intermediate data training for containing images that address the same disease types as in the target datasets.

1. Medico Multimedia Task at MediaEval 2018 (Medico 2018) \citep{jha2021comprehensive} consists of 5,293 GI images, covering 16 classes that include anatomical landmarks, pathological findings, polyp removal cases, and normal images. This dataset aligns with some of the diseases found in the Kvasirv2 dataset, though it spans a wider range of GI images.

2. \href{https://odir2019.grand-challenge.org}{ODIR-2019} (Peking University International Competition on Ocular Disease Intelligent Recognition) stands as an extensive ophthalmic database encompassing 14,400 vividly coloured images captured from both the left and right eyes. Accompanied by diagnostic keywords provided by medical professionals, the dataset spans eight distinct classes, effectively characterising various diseases and conditions depicted in the images. While it shares disease types with the EyePacs dataset, ODIR-2019 is drawn from a different source.

\subsection{Experimental Setup}

\begin{table}[h!]
\centering
\begin{adjustbox}{width=0.95\textwidth}
\begin{tabular}{c c c c}
\hline
\textbf{Dataset} & \textbf{Method} & \textbf{Category} & \textbf{Sample Number} \\
\hline
\multirow{3}{4em}{BusI} & DCGAN & Train Set& 1000 images each class\\
 & - & Test Set& 40 images each class\\
\cmidrule(lr){4-4}
& & & Total: 3120 images \\
\hline
\multirow{3}{4em}{Chest CT } & DCGAN & Train Set& 500 images each class\\
& - & Test Set& 45 images each class\\
\cmidrule(lr){4-4}
& & & Total: 2180 images \\
\hline
\multirow{3}{4em}{Kvasirv2 } & DCGAN & Train Set& 1600 images each class\\
 & - & Test Set& 200 images each class\\
\cmidrule(lr){4-4}
& & & Total: 14400 images \\
\hline
\multirow{3}{4em}{EyePacs } & DCGAN & Train Set& 5000 images each class\\
 & - & Test Set& 500 images each class\\
\cmidrule(lr){4-4}
& & & Total: 11000 images \\
\hline
\end{tabular}
\end{adjustbox}
\caption{The generated target datasets expanded the training sets to 2-4 times the size of the original datasets.}
\label{GAN generated datasets size}
\end{table}

The proposed approach is run in Python 3.10.12 on a Windows 11 pro system under a desktop PC equipped with an Intel (R) Core (TM) i5-12600K CPU at 4.8 GHz, an NVIDIA Geforce RTX 4080 GPU, and 64 GB of RAM. Two pre-trained Xception models initialised with ImageNet weights that pre-trained with TL and SSL methods serve as backbone models for subsequent training. For the base models, the preprocessing and hyperparameter setup mirror those in the previous comparative experiment: all input data are resized to 224x224 pixels. The base model uses the Adam optimizer with an initial learning rate of 0.001. Each training group uses a batch size of 64, with a total of 50 epochs. A learning rate annealer is applied to reduce the learning rate by 0.1 if test loss does not improve after three epochs. For the DCGAN model, the noise dimension is set to 100 with a generation seed of 40, and training uses the Adam optimizer with a learning rate of 0.0002. The batch size for training samples is 16, with a total of 25 epochs. Detailed sample numbers for the generated target dataset are provided in Fig.~\ref{GAN generated datasets size}.

\subsection{Experimental Evaluation}

\begin{figure*}[h!]
\begin{subfigure}{0.5\textwidth}
\includegraphics[width=0.8\linewidth, height=5cm]{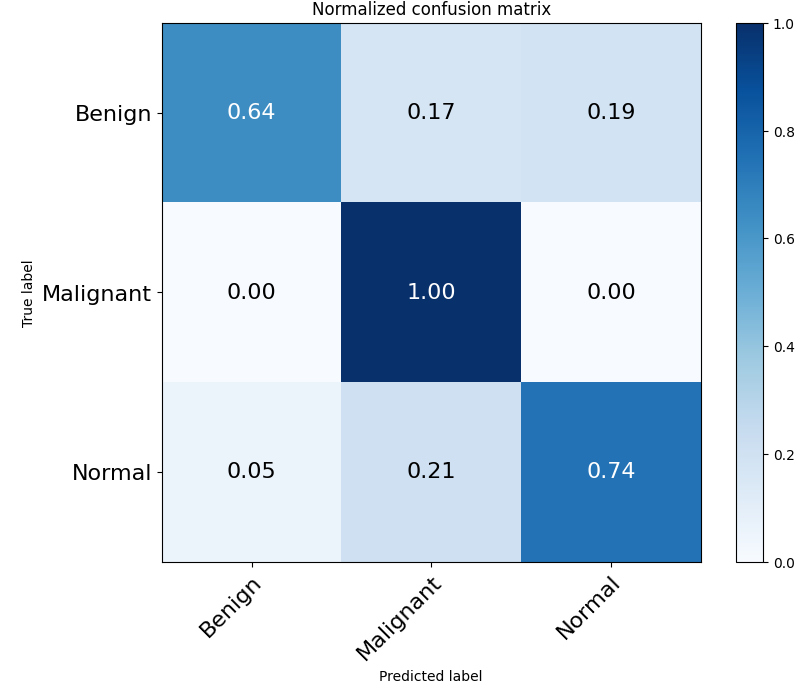} 
\caption{Model trained with only SSL method on BusI dataset}
\label{Base_BusI_SSL}
\end{subfigure}
\begin{subfigure}{0.5\textwidth}
\includegraphics[width=0.8\linewidth, height=5cm]{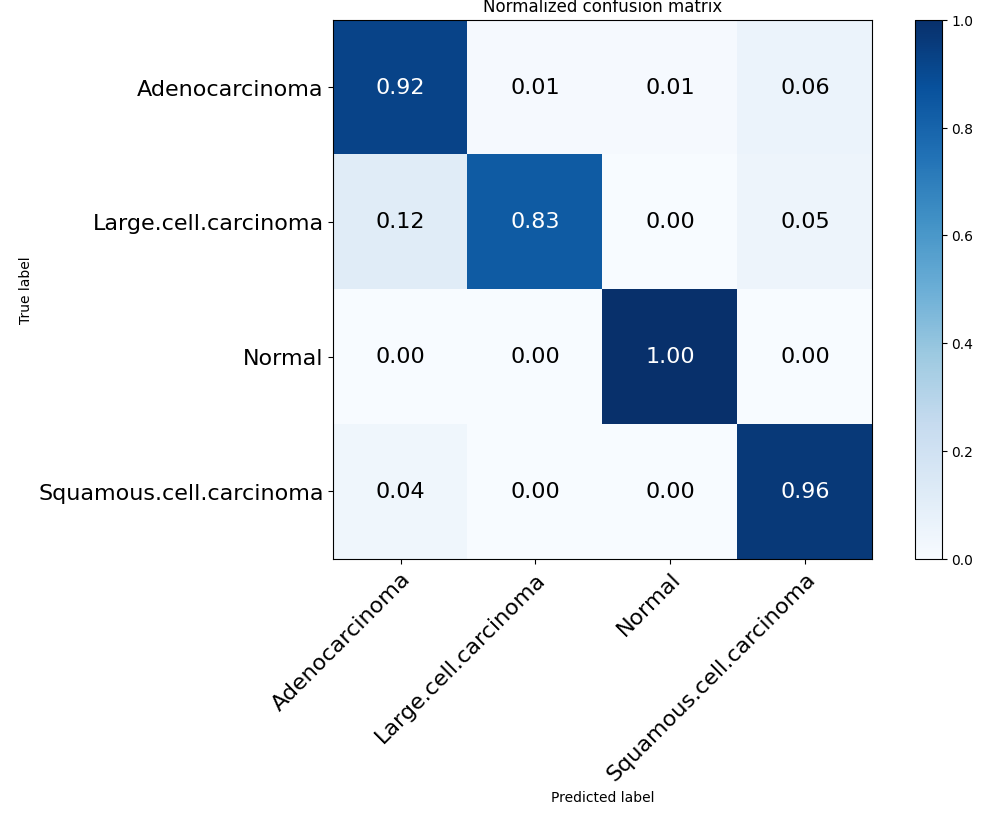}
\caption{Model trained with only SSL method on ChestCT dataset}
\label{Base_ChestCT_SSL}
\end{subfigure}
\begin{subfigure}{0.5\textwidth}
\includegraphics[width=0.8\linewidth, height=5cm]{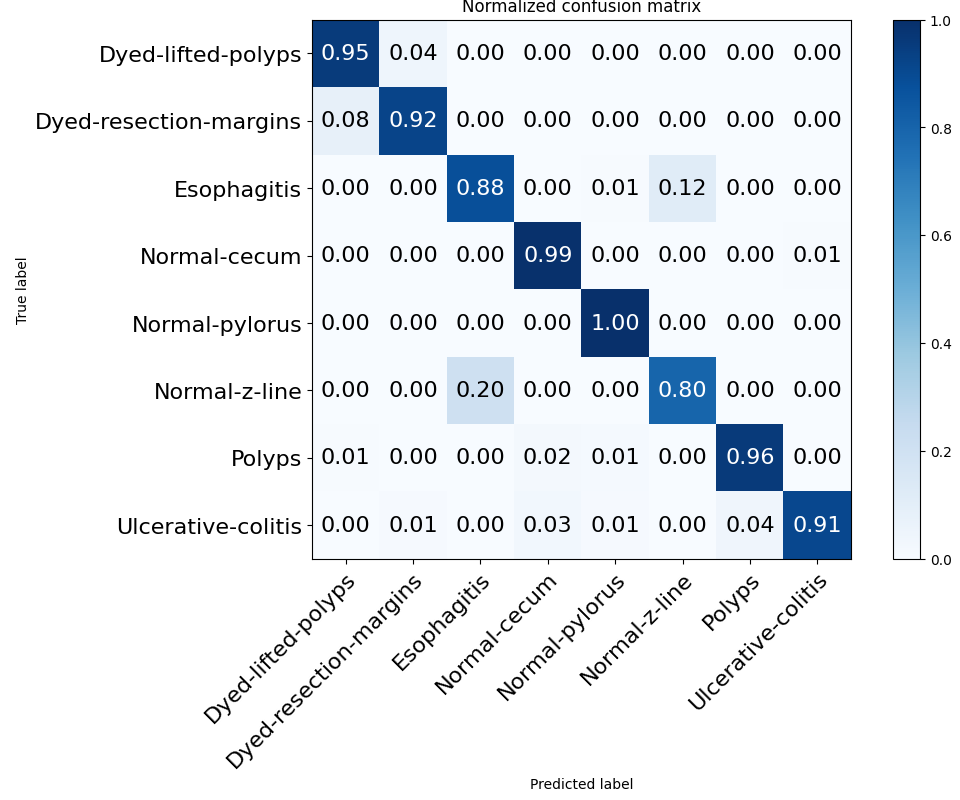}
\caption{Model trained with only TL method on Kavasirv2 dataset}
\label{Base_Kvasir_TL}
\end{subfigure}
\begin{subfigure}{0.5\textwidth}
\includegraphics[width=0.8\linewidth, height=5cm]{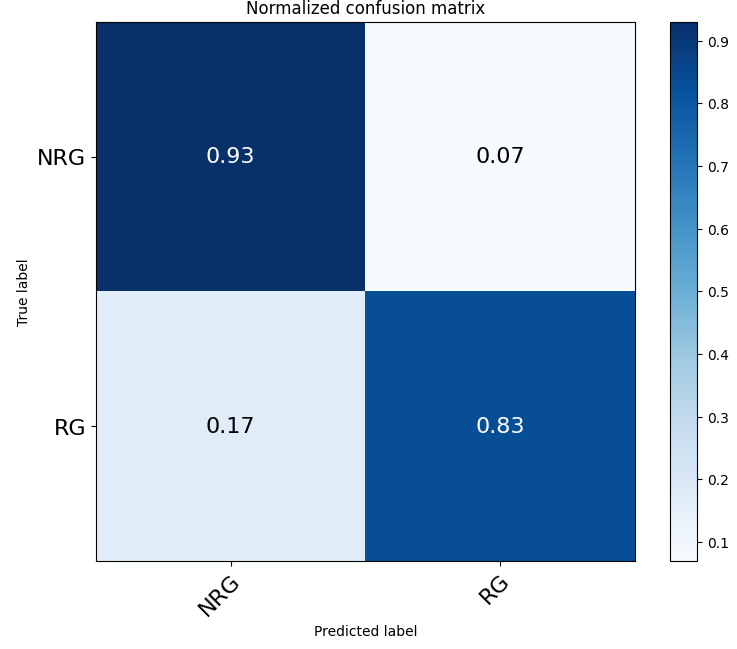} 
\caption{Model trained with only TL method on EyePacs dataset}
\label{Base_EyePacs_TL}
\end{subfigure}
\caption{Confusion matrix (a), (b) represent test result of pre-trained baseline SSL model on BusI and ChestCT dataset, Confusion matrix (c), (d) represent test result of pre-trained baseline TL model on Kvasirv2 and EyePacs dataset.}
\label{Baseline_metrics}
\end{figure*}

\begin{figure*}[h!]
\begin{subfigure}{0.5\textwidth}
\includegraphics[width=0.8\linewidth, height=5cm]{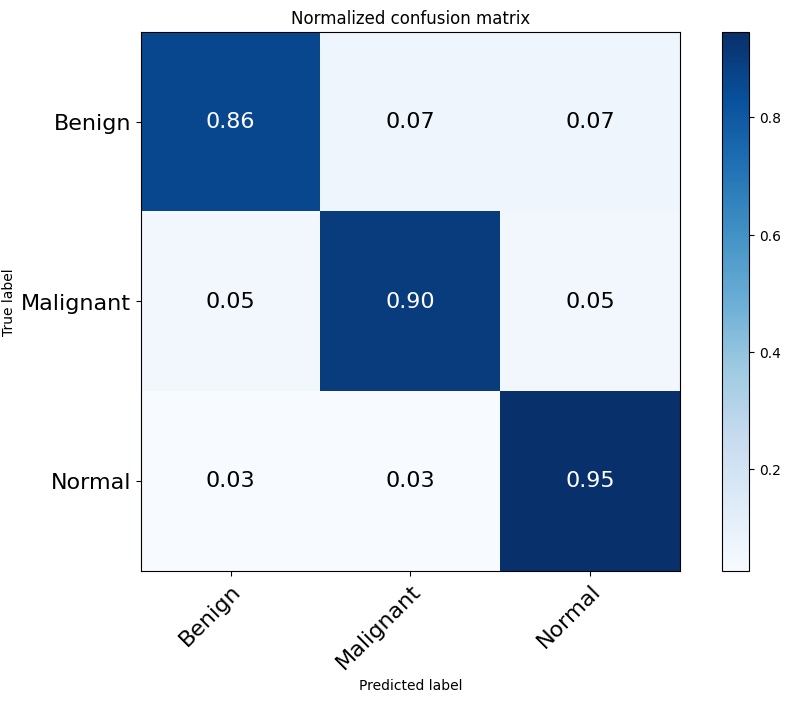} 
\caption{Model trained with proposed approach + SSL method on BusI dataset}
\label{BusI_SSL}
\end{subfigure}
\begin{subfigure}{0.5\textwidth}
\includegraphics[width=0.8\linewidth, height=5cm]{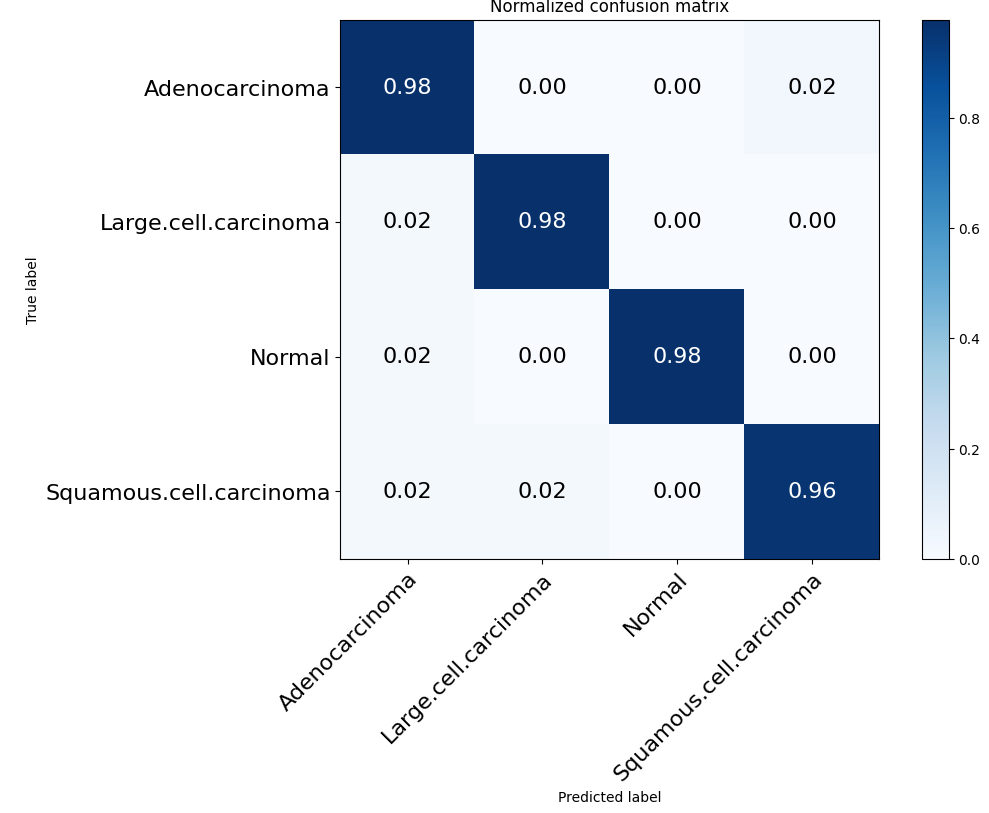}
\caption{Model trained with proposed approach + SSL method on ChestCT dataset}
\label{ChestCT_SSL}
\end{subfigure}
\begin{subfigure}{0.5\textwidth}
\includegraphics[width=0.8\linewidth, height=5cm]{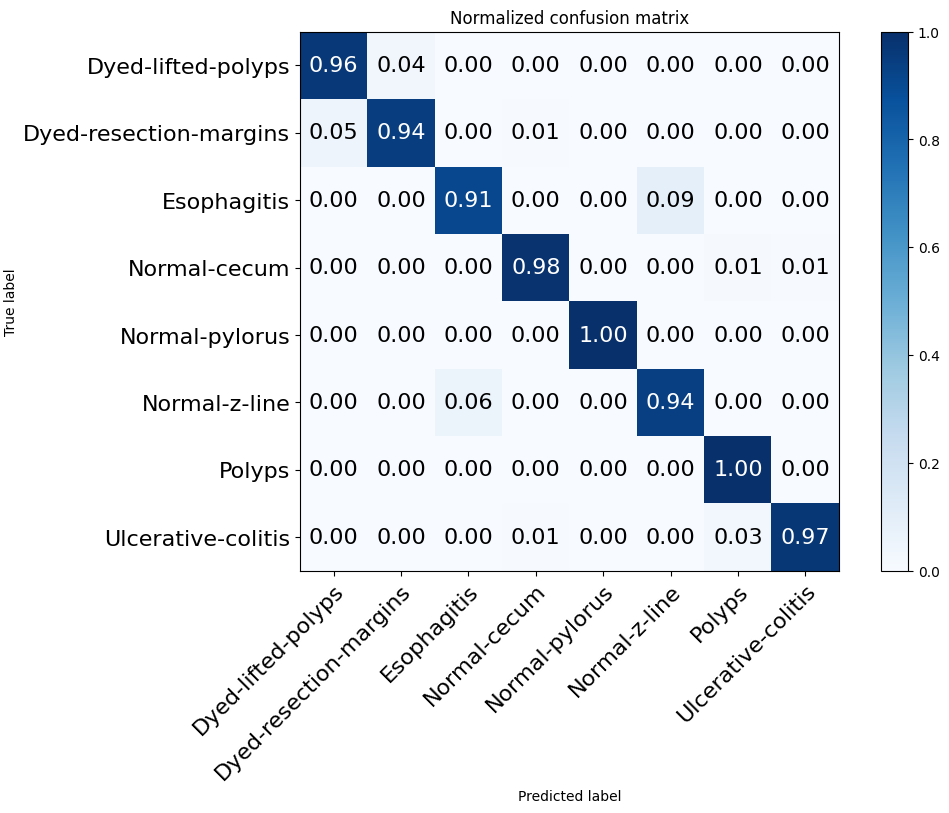}
\caption{Model trained with proposed approach + TL method on Kavasirv2 dataset}
\label{Kvasir_TL}
\end{subfigure}
\begin{subfigure}{0.5\textwidth}
\includegraphics[width=0.8\linewidth, height=5cm]{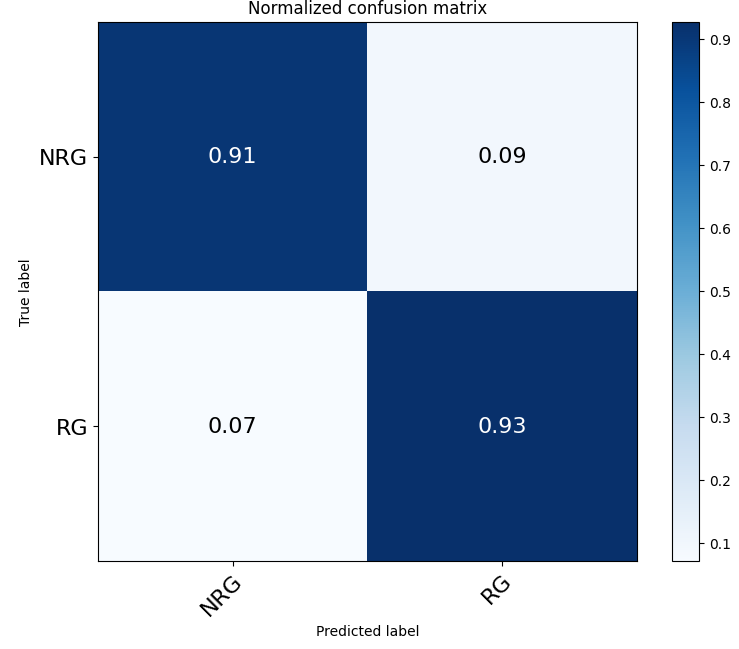} 
\caption{Model trained with proposed approach + TL method on EyePacs dataset}
\label{EyePacs_TL}
\end{subfigure}
\caption{Confusion matrix (a), (b) represent test result of double fine-tuned SSL model on BusI and ChestCT dataset, Confusion matrix (c), (d) represent test result of double fine-tuned TL model on Kvasirv2 and EyePacs dataset. The selected models have achieved the highest test performance on the corresponding target datasets.}
\label{Proposed_Approach_metrics}
\end{figure*}

This section evaluates the test results of the fine-tuned model on the target domain. The test results of each trained model are evaluated using the same evaluation metrics that we used in the comparative experiment and are reported in the following Fig.~\ref{Proposed approach result}, along with the confusion matrix Fig.~\ref{Baseline_metrics} and \ref{Proposed_Approach_metrics} showing specific performance among each class.  

\begin{table*}[h!]
\centering
\begin{adjustbox}{width=1\textwidth}
\begin{tabular}{|c |c |c c c c c |}
\hline
\textbf{Target Dataset} & \textbf{Intermediate Dataset} &\textbf{Method} &\textbf{Acc(\%)}  & \textbf{Sen(\%)} & \textbf{Pre(\%)} & \textbf{F1(\%)}\\
\hline
\multirow{4}{4em}{BusI}  & - & TL& 69.20  & 69.16 & 69.21 & 69.28\\
  & POCS & Proposed+TL& 85.17  & 85.17 & 85.17 & 85.17 \\ 
 & - & SSL& 81.22  & 80.76 & 81.56 & 80.87\\
 & POCS & Proposed+SSL& \textbf{90.67}  & 90.67 & 90.88 & 90.77 \\

\hline
\multirow{4}{4em}{Chest CT}  & - & TL& 87.30  & 87.02 & 87.02 & 87.01 \\
  & CT\_Kidney & Proposed+TL& 94.44 & 94.57 & 94.64 & 94.61\\
 & - & SSL& 94.41  & 94.18 & 93.03 & 93.42 \\
  & CT\_Kidney & Proposed+SSL & \textbf{97.22}  & 97.21 & 97.22 & 97.20 \\
 
\hline
\multirow{4}{4em}{Kvasirv2} & - & TL& 91.50 & 90.90 & 91.00 & 90.87 \\
  & Medico2018 & Proposed+TL& \textbf{96.40}  &96.42 & 96.54 &96.45 \\

 & - & SSL& 85.81  &85.81 & 82.96 & 84.36 \\
  & Medico2018 & Proposed+SSL& 91.73  &91.77 & 91.69 &91.73 \\
\hline
\multirow{4}{4em}{EyePacs} &-& TL&91.20 &91.10 & 91.10&91.00 \\
  &ODIR2019& Proposed+TL& \textbf{92.64}  &91.77 & 92.14 & 91.95 \\
 & - & SSL& 83.20  &83.20& 83.20&83.20 \\
  & ODIR2019 & Proposed+SSL& 90.29 & 91.10& 90.78& 90.94\\
\hline
\end{tabular}
\end{adjustbox}
\caption {The base models pre-trained using the proposed approach, along with TL and SSL methods, are presented separately. Test results for models pre-trained using only the TL or SSL method are also included here as a baseline for comparison with the proposed approach. }
\label{Proposed approach result}
\end{table*}

To better understand performance improvement with the proposed approach, we reused the test results from previous comparative experiments as a baseline for easy comparison. Notably, the pre-trained model we selected as the baseline is the Xception model trained with only the target dataset without using any other data augmentation techniques. As Fig.~\ref{Proposed approach result} shows, the proposed approach combined with two pre-train methods has significantly improved the pre-trained models' performance. The TL-based models have an accuracy improvement ranging from 1.44\% to 17.14\& compared to the baseline models, while SSL-based models also improved from 2.79\% to 9.45\%. Although the performance gap between TL and SSL-based model still exists, the proposed approach has reinforced their performance towards all target datasets. The proposed approach combined with the TL method has yelled a superior model performance when compared to the baseline SSL model, and the same situation occurred with the SSL side. More importantly, the performance gap between the two pre-train methods-based models was also reduced due to the efforts of the intermediate source domain that migrated the domain discrepancies. These results show that the proposed approach effectively addressed the domain mismatch and colour channel discrepancy issues.

Using the DCGAN model for data generation has led to noticeable improvements in model performance compared to results obtained with traditional data augmentation techniques. This demonstrates its potential to help address data scarcity when combined with pre-train methods. At the same time, through Fig.~\ref{Proposed_Approach_metrics} and Fig.~\ref{Baseline_metrics}, we compared the specific prediction results between the baseline model and the proposed approach model. The data imbalance issue influenced the baseline model's prediction results, as the model tended to predict test samples into the class with more training samples. The proposed approach model exhibited more balanced prediction results within the same target datasets, indicating that the bias between prediction results is reduced and that the proposed approach can reduce the influence of data imbalance issues. 

\begin{table*}[h!]
\centering
\begin{adjustbox}{width=1\textwidth}
\begin{tabular}{|c| c c c c c|}
\hline
\textbf{Target Dataset} & \textbf{Source} & \textbf{Model} & \textbf{Pretrain method}  &\textbf{Acc(\%)} & \textbf{F1(\%)} \\
\hline
\multirow{5}{4em}{BusI} &\cite{gheflati2022vision} & ResNet-50 & Transfer learning  & 83.00 & - \\
&\cite{deb2023breast} & Ensemble & Transfer learning  & 85.23 & - \\
& \cite{yue2024medmamba} & MedMamba & Transfer learning & 87.20 & - \\
&Proposed + TL method & Xception & Transfer learning & 85.17 & 85.17\\
&Proposed + SSL method & Xception & Self-supervised learning  & \textbf{90.67} & 90.77 \\
\hline
\multirow{4}{4em}{ChestCT}&\cite{dadgar2022comparative} & InceptionResNetV2 & Transfer learning  & 91.10 & -  \\
&\cite{mamun2023lcdctcnn} & Modified-CNN & -  & 92.00 &  - \\
&Proposed + TL method & Xception & Transfer learning & 94.44 & 94.61 \\
&Proposed + SSL method & Xception & Self-supervised learning &  \textbf{97.22} & 97.20 \\
\hline
\multirow{6}{4em}{Kavasirv2}&\cite{dong2023local} & Ensemble & Transfer learning &  90.60 & 90.70\\
&\cite{mukhtorov2023endoscopic} &ResNet-152 & - & 93.46 & -\\
&\cite{wang2023vision} & Hybrid VIT & - &  95.42 & 95.40\\
&\cite{patel2024classification} & EfficientNetB5 & Transfer learning & 92.60 & 93.00\\
&Proposed + TL method & Xception & Transfer learning &  \textbf{96.40} & 96.45 \\
&Proposed + SSL method & Xception & Self-supervised learning &  91.73 & 91.73 \\
\hline
\multirow{3}{4em}{EyePacs}&\cite{kiefer2023automated} & MobileNetV3 & Transfer Learning &  88.30 & -\\
&Proposed TL method & Xception & Transfer learning &  \textbf{92.64} & 91.95 \\
&Proposed SSL method & Xception & Self-supervised learning & 90.29 & 90.94\\
\hline
\end{tabular}
\end{adjustbox}
\caption{The model performance comparison between the proposed approach and state-of-the-art works.}
\label{SOTA model comparing}
\end{table*}

Additionally, we compared the proposed approach with state-of-the-art works. The objective is to elucidate the efficacy of the proposed pre-trained models on the selected datasets, and the results are delineated in Table~\ref{SOTA model comparing}. A discerning examination of the table reveals the superior performance of our proposed TL and SSL models compared to state-of-the-art models on four datasets. The proposed SSL model has achieved advantages of 4.8\% and 5.2\% in accuracy compared to those state-of-the-art DL models that were either trained from scratch or employed transfer learning methodologies. This outcome aligns with the observation of earlier comparative experiments, indicating that the SSL method has more advantages in learning greyscale medical images than the TL method. Concurrently, the proposed TL model demonstrated competitive prowess when compared to state-of-the-art DL models, showing a 1\% improvement in the KvasirV2 dataset and a 4\% improvement in the EyePacs dataset. This observation underscores the improved generalisation ability and advanced performance of the proposed approach combined with TL methods.

Furthermore, comparing with work trained from scratch, the proposed TL model has shown additional advantages in training speed. For instance, \cite{mukhtorov2023endoscopic} reported that their CNN model, developed without the use of pre-training, necessitated a training duration of 6-12 hours from inception on the target Kvasirv2 dataset. In stark contrast, the pre-trained TL model required a mere 40 minutes for fine-tuning on both the intermediate and target datasets, concurrently maintaining a better performance level during the testing phase. However, on average, the pre-trained SSL model required double or triple the time compared to the TL model, consuming more time and computational resources in the training process. This high time consumption of SSL training should be paid more attention to achieve a balance between model performance and reasonable time cost.

\subsection{Explainable Evaluation}

The final step involves using the explainable AI technique to analyse the proposed approach's robustness, elucidate the trained model's decision-making process, and assess its reliability. As emphasised in the literature review and recent studies \citep{albahri2023systematic, quinn2022three}, the reliability of AI applications in bio-medicine and healthcare has become a focal point of concern among researchers and medical institutions. The trustworthiness holds pivotal significance for AI's widespread acceptance and integration within the medical domain. Feature visualisation is valuable for uncovering and understanding the learnt features within DL models. \cite{zhou2016learning} introduced the Class Activation Mapping (CAM) method, which incorporates a global average pooling layer into a standard CNN. This innovation facilitated the identification of critical feature contributions linked to CNN's specific predictions, shedding light on the rationale behind the model's decisions. Furthermore, \cite{selvaraju2017grad} introduced Gradient Weighted Class Activation Mapping (Grad-CAM), using gradients from the network's final convolutional layer to generate a coarse localisation map. This map highlights influential regions within an image that contribute to the prediction of specific concepts or classes. In our study, we apply Grad-CAM on the model’s final layer to generate class-specific heat maps, revealing which regions contribute to the model’s prediction rationale. Specifically, we visualised predictions from baseline models, focusing on misclassified samples, then compared them to those from the proposed approach model. This visual comparison between baselines and proposed approach offers insights into model reliability and highlights areas for improvement in incorrect predictions.

\begin{table*}[h!]
\centering
\begin{adjustbox}{width=1\textwidth}
\begin{tabular}{c c c c}
\hline
\textbf{Dataset} & \textbf{Baeline TL Model} & \textbf{Proposed SSL Model} & \textbf{True Label}  \\
\hline
BusI & Pred: \textcolor{red}{Benign} &  Pred: \textcolor{green}{Malignant} &  Malignant \\
\includegraphics[width=0.3\linewidth, height=2.8cm]{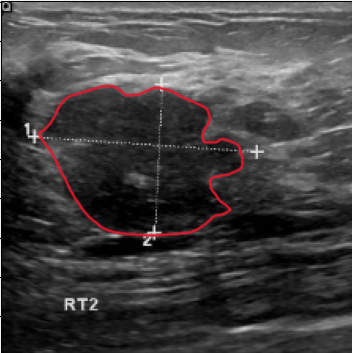} 
& \includegraphics[width=0.3\linewidth, height=2.8cm]{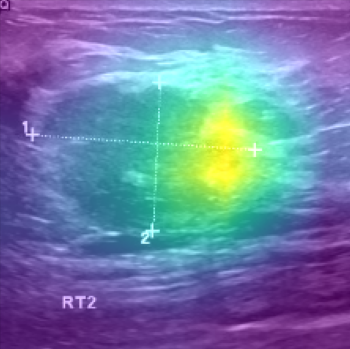} 
& \includegraphics[width=0.3\linewidth, height=2.8cm]
{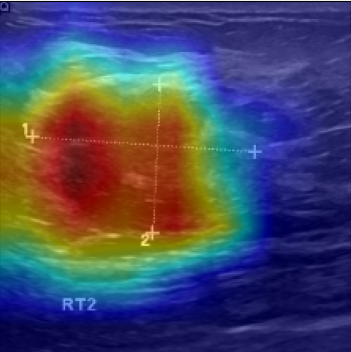} &\\
\hline
BusI  & Pred: \textcolor{red}{Benign} &  Pred: \textcolor{green}{Malignant} &  Malignant \\
\includegraphics[width=0.3\linewidth, height=2.8cm]{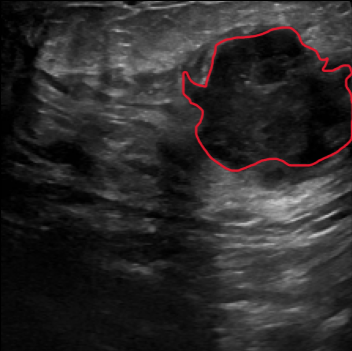} 
& \includegraphics[width=0.3\linewidth, height=2.8cm]{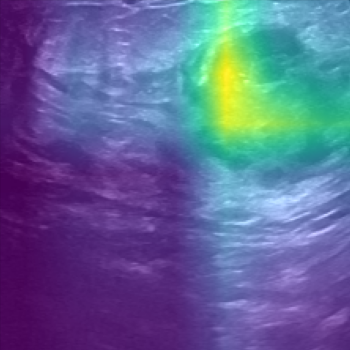} 
& \includegraphics[width=0.3\linewidth, height=2.8cm]
{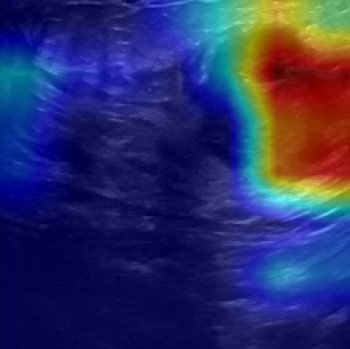} &\\
\hline
BusI  & Pred: \textcolor{red}{Benign} &  Pred: \textcolor{green}{Malignant} &  Malignant \\
\includegraphics[width=0.3\linewidth, height=2.8cm]{BusI_GradCAM_Heatmap/Breast_Base_Downsample_original.png} 
& \includegraphics[width=0.3\linewidth, height=2.8cm]{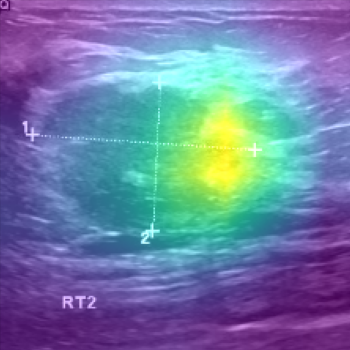} 
& \includegraphics[width=0.3\linewidth, height=2.8cm]
{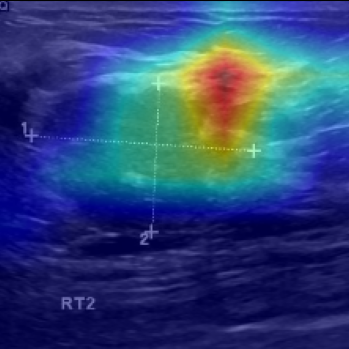} &\\
\hline
ChestCT & Pred: \textcolor{red}{Adenocarcinoma} &  Pred: \textcolor{green}{Squamous.Cell} &  Squamous.Cell \\
\includegraphics[width=0.3\linewidth, height=2.8cm]{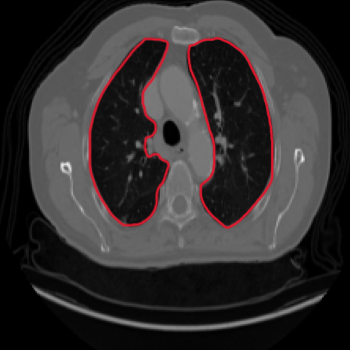} 
& \includegraphics[width=0.3\linewidth, height=2.8cm]{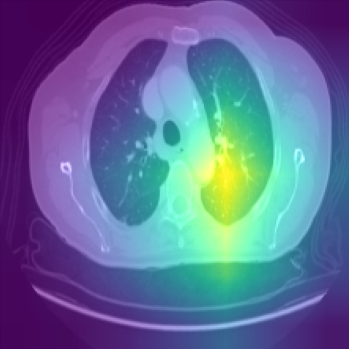} 
& \includegraphics[width=0.3\linewidth, height=2.8cm]
{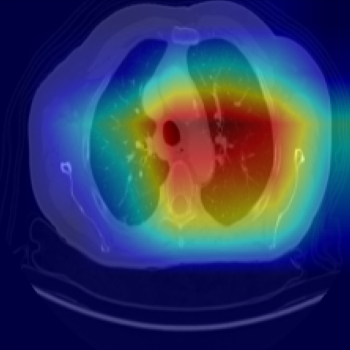} &\\
\hline
ChestCT & Pred: \textcolor{red}{Adenocarcinoma} &  Pred: \textcolor{green}{Large.Cell} &  Large.Cell \\
\includegraphics[width=0.3\linewidth, height=2.8cm]{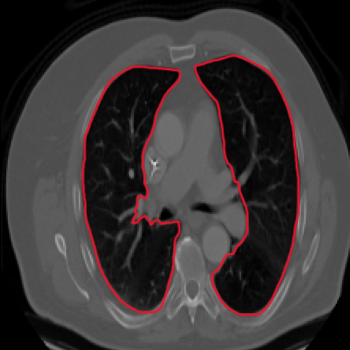} 
& \includegraphics[width=0.3\linewidth, height=2.8cm]{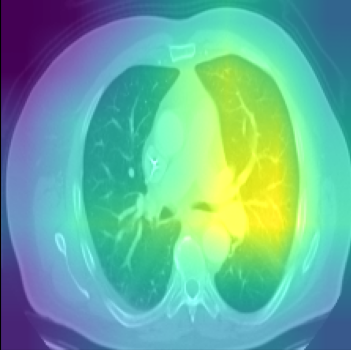} 
& \includegraphics[width=0.3\linewidth, height=2.8cm]
{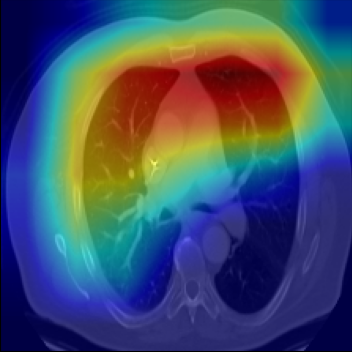} &\\
\hline
ChestCT  & Pred: \textcolor{red}{Squamous.Cell} &  Pred: \textcolor{green}{Large.Cell} &  Large.Cell \\
\includegraphics[width=0.3\linewidth, height=2.8cm]{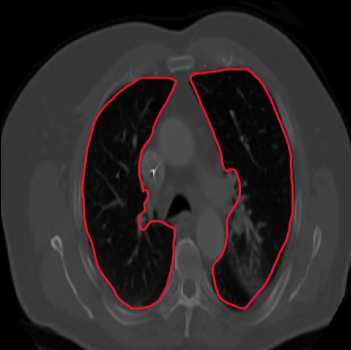} 
& \includegraphics[width=0.3\linewidth, height=2.8cm]{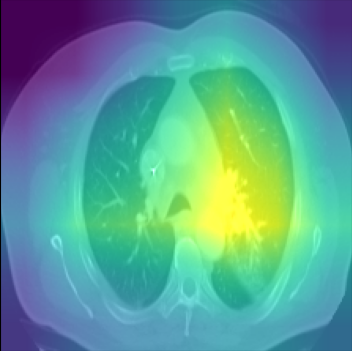} 
& \includegraphics[width=0.3\linewidth, height=2.8cm]
{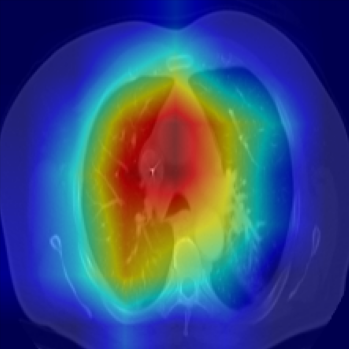} &\\
\hline
\end{tabular}
\end{adjustbox}
\caption{The comparison of baseline TL and proposed SSL models' visualisation heatmaps towards grey-scale target datasets.}
\label{Robustness comparison on grey-scale}
\end{table*}

\begin{table*}[h!]
\centering
\begin{adjustbox}{width=1\textwidth}
\begin{tabular}{c c c c}
\hline
\textbf{Datasets} & \textbf{Proposed TL Model} & \textbf{Baseline SSL Model} & \textbf{True Label}  \\
\hline
KvasirV2 & Pred: \textcolor{green}{Polyps} &  Pred: \textcolor{red}{Ulcerative-colitis} &  Polyps \\
\includegraphics[width=0.3\linewidth, height=2.8cm]{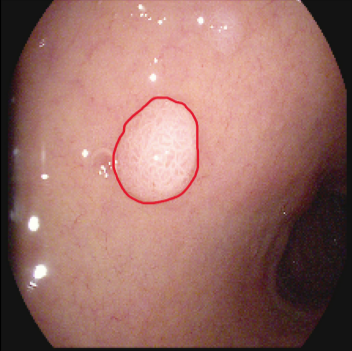} 
& \includegraphics[width=0.3\linewidth, height=2.8cm]{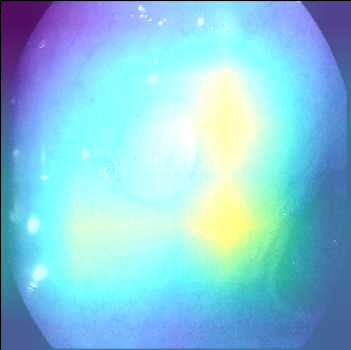} 
& \includegraphics[width=0.3\linewidth, height=2.8cm]
{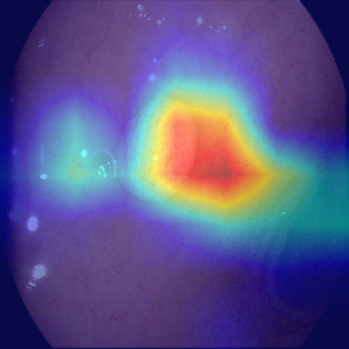} &\\
\hline
KvasirV2 & Pred: \textcolor{green}{Normal-z-line} &  Pred: \textcolor{red}{Normal-pylorus} &  Normal-z-line \\
\includegraphics[width=0.3\linewidth, height=2.8cm]{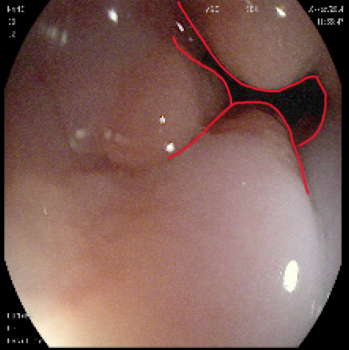} 
& \includegraphics[width=0.3\linewidth, height=2.8cm]{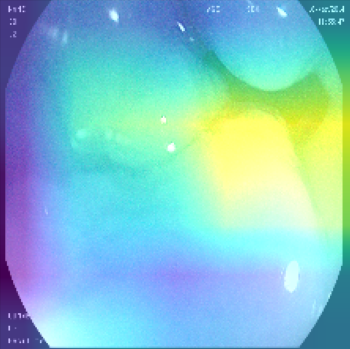} 
& \includegraphics[width=0.3\linewidth, height=2.8cm]
{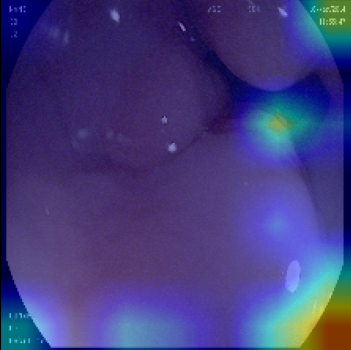} &\\
\hline
KvasirV2 & Pred: \textcolor{red}{Ulcerative-colitis} &  Pred: \textcolor{green}{Polyps} &  Polyps \\
\includegraphics[width=0.3\linewidth, height=2.8cm]{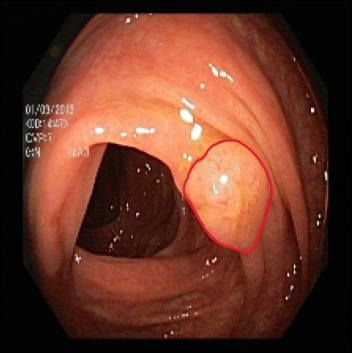} 
& \includegraphics[width=0.3\linewidth, height=2.8cm]{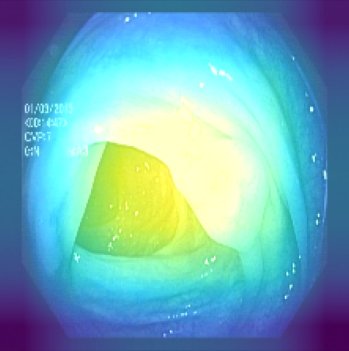} 
& \includegraphics[width=0.3\linewidth, height=2.8cm]
{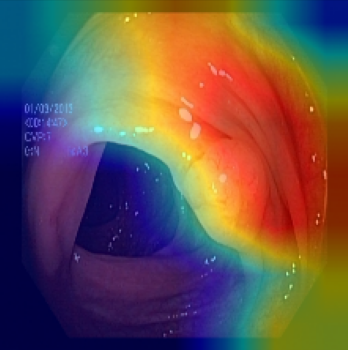} &\\
\hline
EyePacs & Pred: \textcolor{green}{RG} &  Pred: \textcolor{red}{NRG} &  RG \\
\includegraphics[width=0.3\linewidth, height=2.8cm]{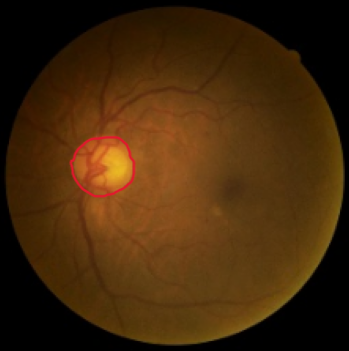} 
& \includegraphics[width=0.3\linewidth, height=2.8cm]{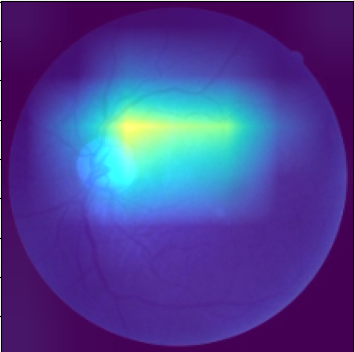} 
& \includegraphics[width=0.3\linewidth, height=2.8cm]
{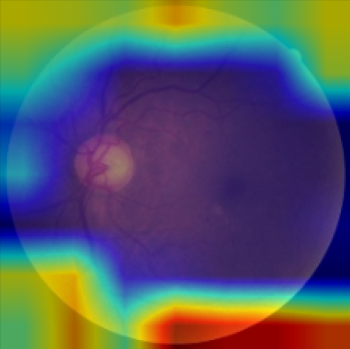} &\\
\hline
EyePacs & Pred: \textcolor{green}{RG} &  Pred: \textcolor{red}{NRG} &  RG \\
\includegraphics[width=0.3\linewidth, height=2.8cm]{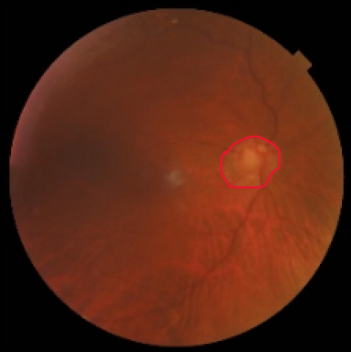} 
& \includegraphics[width=0.3\linewidth, height=2.8cm]{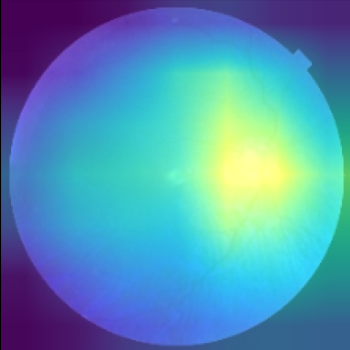} 
& \includegraphics[width=0.3\linewidth, height=2.8cm]
{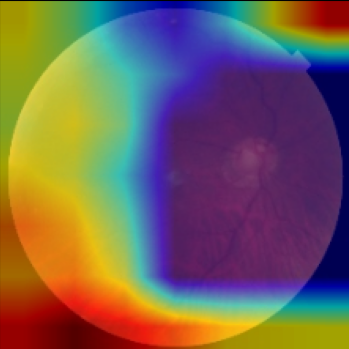} &\\
\hline
EyePacs & Pred: \textcolor{green}{NRG} &  Pred: \textcolor{red}{RG} &  NRG \\
\includegraphics[width=0.3\linewidth, height=2.8cm]{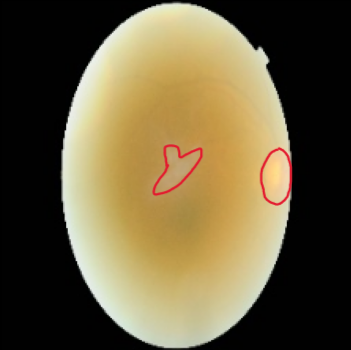} 
& \includegraphics[width=0.3\linewidth, height=2.8cm]{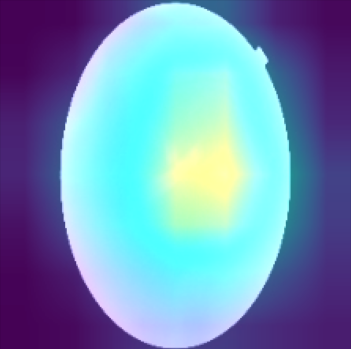} 
& \includegraphics[width=0.3\linewidth, height=2.8cm]
{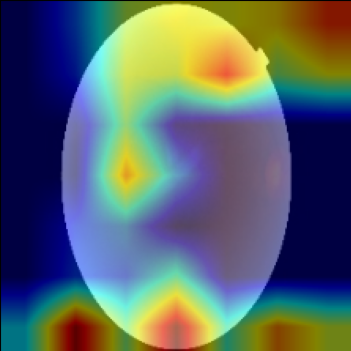} &\\
\hline
\end{tabular}
\end{adjustbox}
\caption{The comparison of proposed TL and baseline SSL models' visualisation heatmaps towards colourful target datasets.}
\label{Robustness comparison on colourful}
\end{table*}

In Table~\ref{Robustness comparison on grey-scale}, we conducted a comparative analysis of heatmaps generated from predicted samples sourced from grey-scale datasets including BusI and Chest CT. Each set includes the original sample with marked disease regions, followed by two heatmaps predicted by the baseline TL and proposed SSL models, respectively. For the BusI dataset, we focused on samples from the Malignant class, as the primary objective of this dataset is to distinguish malignant tumours that pose a threat to patient health. Malignant cancer typically presents with large tumour segments, shadows, and multiple projections from nodules. In the first BusI dataset sample, the baseline TL model misclassified it as Benign, focusing only on a fraction of the disease region, failing to encompass the entire area. In contrast, the proposed SSL model correctly identified the disease region, indicating its ability to discern malignant tumours accurately. This trend persisted in the second sample, where the TL model again failed to fully capture the disease region, while the SSL model made the correct prediction. However, in the third sample, both the TL and SSL model lost focus on the true disease region, even if the SSL model have made a correct prediction. This highlights the potential risk of SSL method in predicting results based on unrelated feature representations, which is caused by lacking strong supervision signals during the training process.

For the Chest CT dataset, samples were selected from the Squamous.cell.carcinoma and Large.cell.carcinoma classes, representing the main symptoms of lung cancer diseases. Squamous.cell.carcinoma typically exhibits cavitation and lung collapse, while Large.cell.carcinoma presents as large peripheral masses with irregular margins in lung segments. Accordingly, the trained models are supposed to focus on the lung sides rather than soft tissues as it shown in the sample masks. For the first image sample, the TL model misclassified the disease as Adenocarcinoma, focusing its attention on the bottom segment of the right lung. In contrast, the SSL model correctly classified the sample, directing its attention to the right lung side and middle aorta and cavity, aligning more closely with the symptoms. In the second sample from the Large.cell.carcinoma class, the TL model improved its focus but still mispredicted the label, whereas the SSL model correctly classified the sample by widening its focus to both lung sides and middle tissue. In the third ChestCT sample, the TL model remained focusing on the right lung segment, put its attention on white bubbles and nerves at the bottom, resulting in an incorrect prediction of Squamous.cell.carcinoma. In contrast, the SSL model correctly focused on the entire left lung segment and middle cavity, leading to an accurate prediction. The visulisation in the ChestCT samples generally shows that the proposed SSL model can successfully learned and captured the target disease region, while the baseline TL model did not fully understand the sample due to the lack of training samples. 

Table~\ref{Robustness comparison on colourful} presents a series of heat maps generated from colourful target datasets, specifically KvasirV2 and EyePacs. Within the KvasirV2 dataset, our analysis revealed challenges in classifying three pairs of classes: Dyed-lifted polyps versus Dyed-resection-margins, Esophagitis versus Normal z-line, Polyps versus Ulcerative-colitis. Among these pairs, Esophagitis and Polyps represent common esoph-ageal diseases, prompting our focus on samples from the normal z-line and polyps to discern how TL and SSL models differentiate between them. The normal z-line typically features a clear demarcation at the oesophagus-stomach junction with uniform colour and smooth texture, similar to surrounding tissue. In contrast, Esophagitis often manifests as inflammation, erosion, or ulceration along the esophageal lining. Polyps exhibit discrete raised lesions with a smooth or lobulated surface, while Ulcerative Colitis presents diffuse inflammation, erythema, loss of vascular pattern, and friability in a continuous pattern. Taking advantage of these distinctions, we annotated the disease regions in the original images for comparative analysis. In the KvasirV2 dataset, the first Polyps sample shows substantial caking on the intestinal surface. The TL model correctly directed its attention around the caking, while the SSL model erroneously fixated on the caking itself, resulting in an incorrect prediction. In another instance from the KvasirV2 dataset, the TL model effectively discerned the demarcation between tissue and texture, accurately classifying the sample. Conversely, the SSL model misidentified the background, leading to an incorrect prediction of Normal-pylorus. In the third KvasirV2 sample, the attention of the TL model was dispersed between intestinal pores and caking tissue, resulting in misclassification. In contrast, the SSL model accurately identified Polyps by focusing on the caking tissue and part of the intestinal lining.

In the EyePacs dataset, the task revolves around distinguishing referable glaucoma (RG) and non-referable glaucoma (NRG) from colourful fundus images. Samples from both RG and NRG classes were examined. RG samples typically exhibit pronounced optic disc features such as cup-to-disc ratio, neuroretinal rim thinning, disc hemorrhages, and notching compared to NRG. The TL model effectively directed attention to the disc area, correctly classifying RG images. In contrast, the SSL model was fixed on the background of the image, leading to misclassifications. In the second sample, the TL model maintained accurate predictions by focusing on the disc area, while the SSL model failed to improve its performance, and fixed on the background. In the last sample with few optic disc features, the TL model correctly identified the white representation area, while the attention of the SSL model was divided between the eyecup and background, indicating a lack of true feature comprehension in NRG representations.

In summary, the heatmap samples generated through the Grad-CAM technique show that the proposed approach combined TL and SSL models not only improved the performance of the baseline models but also gained more understanding of the target samples and made predictions more robust. These findings highlight the efficacy of the proposed approach in addressing the related issues within the medical field and improving the pre-train methods.

\section{Discussion}

In this section, we discuss our findings from the experiments and offer guidance for researchers aiming to apply pre-train methods in their work. First, we explored the strengths and limitations of transfer learning and self-supervised learning methods by conducting a comparative experiment between them. This analysis highlights that the performance gap between these two methods is influenced by multiple factors, including the size of the training dataset, the class imbalance, the differences in colour information, and the distinct advantages of TL and SSL for handling various types of data. Specifically, our experiments reveal that TL, particularly with ImageNet pre-trained weights, performs well on colourful medical datasets but is less effective on grey-scale datasets. In contrast, SSL shows higher performance with greyscale datasets, but is less effective for colourful datasets. Although both methods aim to address data scarcity, SSL models generally exhibit more stability with smaller datasets than TL models. We also examined the effectiveness of data augmentation techniques combined with pre-trained models in addressing data imbalance issues. Our results show that general augmentation techniques benefit the fine-tuning process while combining a generative model, such as the DCGAN model, with pre-training further boosts model performance and enhance robustness. This suggests that data augmentation is critical to maximising the success of pre-trained models.

Another key challenge discussed in this study is the lack of a medical-specific source domain. Although SSL’s unsupervised approach mitigates this issue to some extent, challenges remain, such as limited supervision during pre-training or the absence of relevant samples for rare diseases. Instead, the proposed approach using double fine-tuning technique revealed impressive improvement towards both TL and SSL methods. Although the original studies of distance pre-train method are based on TL method, observations affirm the utility of double fine-tuning in enhancing SSL models' performance and robustness at the same time. We believe utilising double fine-tuning technique has strong potential to address many of the challenges in medical AI.

The insights gained in this study can be extended beyond medical research, providing a foundation for addressing data scarcity across various fields through pre-train methods. Additionally, we offer practical guidance for researchers looking to incorporate pre-training into their studies.

\begin{enumerate}

\item We recommend the use of the SSL method in small target datasets (for example: less than 1000 samples), as it imparts superior performance and stability compared to the TL method when facing challenging scenarios.

\item The nature of the colour information within the target data samples dictates the choice between the TL and SSL methods. TL methods excel with colourful images, while SSL methods exhibit proficiency with grey-scale images. Choosing right pre-train methods for the specific target domain can lead to a more efficient training process and better performance.

\item When confronting imbalanced data, it is advisable to employ the data augmentation technique with pre-train methods. However, researchers need to pay more attention to the associated increase in training time and computational resource consumption, as well as to the risk of data duplication.

\item In the absence of extensive source data that mirror target data features, consider the double fine-tuning approach: pre-training the model in a less related domain with more samples, followed by fine-tuning it in a closely related domain with smaller samples. This strategy bridges the gap between pre-trained models and target datasets without requiring abundant similar training samples.

\item Pre-train methods have potential to extend their utility beyond small datasets; Even relatively large datasets benefit from pre-train methods with improved performance, robustness, training speed, and reduced cost of computational resource. Researchers are encouraged to explore pre-train methods within and beyond the medical field.

\end{enumerate}

\section{Conclusion}

This article presents a comparative study of TL and SSL methods to examine the factors that affect their test performance, particularly within medical applications. The experiment focused on three key issues in this domain: data scarcity, data imbalance, and data discrepancy, analysing their impact on the effectiveness of pre-train methods. To address limitations observed in these methods' reliability and applicability, we proposed a hybrid approach that combines double fine-tuning with data generation techniques to enhance performance. Evaluation was conducted across four diverse medical datasets including two grey-scale and two colourful ones covering different disease types. The proposed SSL model achieved 90.67\% and 97.22\% accuracy in multiclass classification tasks for the target BusI and Chest CT datasets, while the TL model achieved 96.4\% and 92.64\% accuracy in classifying image classes for the Kvasirv2 and EyePacs datasets. This approach shows promising performance in supporting disease diagnosis and classification when integrated with image-based AI, offering improved robustness in identifying disease regions and making accurate predictions.

Due to the experiment’s scope, we focused on the medical domain and tested only four base architectures. In this study, issues such as label misclassification, human error image noise, and environmental factors such as the effects of weather and lighting on backgrounds were not addressed. Future work aims to test pre-train methods in broader contexts to better understand their performance differences and unique strengths. We also plan to apply our method to more fields where training data is limited, enhancing model performance across diverse applications.

\backmatter

\bmhead{\large{List of abbreviations}}

\begin{flushleft}\textbf{DL}: Deep Learning\end{flushleft}

\begin{flushleft}\textbf{TL}: Transfer Learning\end{flushleft}

\begin{flushleft}\textbf{SSL}: Self-supervised Learning\end{flushleft}

\begin{flushleft}\textbf{DCGAN}: Deep Convolutional Generative Adversary Network\end{flushleft}

\begin{flushleft}\textbf{XAI}: Explainable Artificial Intelligence\end{flushleft}

\begin{flushleft}\textbf{SimCLR}: Simple Framework for Contrastive Learning of Visual \end{flushleft}Representations

\begin{flushleft}\textbf{POCAS}: Polycystic Ovary Syndrome\end{flushleft}

\begin{flushleft}\textbf{CAM}: Class Activation Mapping\end{flushleft}

\begin{flushleft}\textbf{Grad-CAM}: Gradient Weighted Class Activation Mapping\end{flushleft}

\begin{flushleft}\textbf{RG}: Referable Glaucoma\end{flushleft}

\begin{flushleft}\textbf{NRG}: Non-referable Glaucoma\end{flushleft}

\begin{flushleft}\large{\textbf{Funding}}\end{flushleft}

The authors acknowledge the support received through the following funding schemes of the Australian Government: Australian Research Council (ARC) Industrial Transformation Training Centre (ITTC) for Joint Biomechanics under grant (IC190100020). \\

\begin{flushleft}\large{\textbf{Clinical trial number:  Not applicable.} }\end{flushleft}

\begin{flushleft}\large{\textbf{Consent to Publish declaration:  Not applicable.} }\end{flushleft}

\bmhead{\large{Data availability}}

All the datasets used in this study are based on publicly available datasets, including the source, intermediate, and target domain datasets. For the breast ultrasound classification task, the intermediate dataset \href{https://www.kaggle.com/datasets/anaghachoudhari/pcos-detection-using-ultrasound-images}{POCS} and target dataset \href{https://www.kaggle.com/datasets/sabahesaraki/breast-ultrasound-images-dataset}{BusI} are both available from Kaggle library. For the chest cancer classification task, the intermediate dataset \href{https://www.kaggle.com/datasets/nazmul0087/ct-kidney-dataset-normal-cyst-tumor-and-stone?select=CT-KIDNEY-DATASET-Normal-Cyst-Tumor-Stone}{CT-Kidney} and target dataset \href{https://www.kaggle.com/datasets/mohamedhanyyy/chest-ctscan-images}{ChestCT} are from under the open database license and are publicly available from Kaggle library. For the gastrointestinal endoscopy classification task, the intermediate dataset \href{http://www.multimediaeval.org/mediaeval2018/medico/}{Medico2018} is based on a public medical task, and the corresponding data are available to participants and other multimedia researchers without restriction. The target dataset \href{https://www.kaggle.com/datasets/plhalvorsen/kvasir-v2-a-gastrointestinal-tract-dataset}{Kvasirv2} is under the CC BY-SA 4.0 licence and can be accessed from Kaggle library. For the glaucoma classification task, the intermediate dataset \href{https://odir2019.grand-challenge.org}{ODIR-2019} and target dataset \href{https://www.kaggle.com/datasets/deathtrooper/glaucoma-dataset-eyepacs-airogs-light-v2}{EyePacs} are both publicly available datasets share within Kaggle library. Moreover, the source domain dataset ImageNet Large Scale Visual Recognition Challenge (ILSVRC) that has been used for each base model's pre-training is publicly available via \href{https://image-net.org/download.php}{ImageNet Website}.

\begin{flushleft}\large{\textbf{Ethics declaration: Not applicable.} }\end{flushleft}

\begin{flushleft}\large{\textbf{Consent to Publish declaration: Not applicable.} }\end{flushleft}

\begin{flushleft}\large{\textbf{Consent to Participate declaration: Not applicable.} }\end{flushleft}

\begin{flushleft} \large\textbf{{Competing interests }}\end{flushleft}

All authors have declared that no conflict of interest exists. \\

\begin{flushleft} \large\textbf{{Author contribution}}\end{flushleft} 

\textbf{Zehui Zhao:} Conceptualisation, Methodology,  Validation, Writing: original draft, \textbf{Laith Alzubaidi:} Conceptualisation, Methodology, Resources, Formal analysis, Writing: review \& editing, Validation, Supervision, \textbf{Jinglan Zhang:}Writing: review \& editing, Formal analysis, Supervision \textbf{Ye Duan:} Formal analysis, Writing: review \& editing, \textbf{Usman Naseem} Formal analysis, Writing: review \& editing, \textbf{Yuantong Gu:} Writing: review \& editing, Validation, Funding acquisition. All authors reviewed the manuscript. \\

\bibliography{citation_list}


\begin{thebibliography}{61}
\ifx \bisbn   \undefined \def \bisbn  #1{ISBN #1}\fi
\ifx \binits  \undefined \def \binits#1{#1}\fi
\ifx \bauthor  \undefined \def \bauthor#1{#1}\fi
\ifx \batitle  \undefined \def \batitle#1{#1}\fi
\ifx \bjtitle  \undefined \def \bjtitle#1{#1}\fi
\ifx \bvolume  \undefined \def \bvolume#1{\textbf{#1}}\fi
\ifx \byear  \undefined \def \byear#1{#1}\fi
\ifx \bissue  \undefined \def \bissue#1{#1}\fi
\ifx \bfpage  \undefined \def \bfpage#1{#1}\fi
\ifx \blpage  \undefined \def \blpage #1{#1}\fi
\ifx \burl  \undefined \def \burl#1{\textsf{#1}}\fi
\ifx \doiurl  \undefined \def \doiurl#1{\url{https://doi.org/#1}}\fi
\ifx \betal  \undefined \def \betal{\textit{et al.}}\fi
\ifx \binstitute  \undefined \def \binstitute#1{#1}\fi
\ifx \binstitutionaled  \undefined \def \binstitutionaled#1{#1}\fi
\ifx \bctitle  \undefined \def \bctitle#1{#1}\fi
\ifx \beditor  \undefined \def \beditor#1{#1}\fi
\ifx \bpublisher  \undefined \def \bpublisher#1{#1}\fi
\ifx \bbtitle  \undefined \def \bbtitle#1{#1}\fi
\ifx \bedition  \undefined \def \bedition#1{#1}\fi
\ifx \bseriesno  \undefined \def \bseriesno#1{#1}\fi
\ifx \blocation  \undefined \def \blocation#1{#1}\fi
\ifx \bsertitle  \undefined \def \bsertitle#1{#1}\fi
\ifx \bsnm \undefined \def \bsnm#1{#1}\fi
\ifx \bsuffix \undefined \def \bsuffix#1{#1}\fi
\ifx \bparticle \undefined \def \bparticle#1{#1}\fi
\ifx \barticle \undefined \def \barticle#1{#1}\fi
\bibcommenthead
\ifx \bconfdate \undefined \def \bconfdate #1{#1}\fi
\ifx \botherref \undefined \def \botherref #1{#1}\fi
\ifx \url \undefined \def \url#1{\textsf{#1}}\fi
\ifx \bchapter \undefined \def \bchapter#1{#1}\fi
\ifx \bbook \undefined \def \bbook#1{#1}\fi
\ifx \bcomment \undefined \def \bcomment#1{#1}\fi
\ifx \oauthor \undefined \def \oauthor#1{#1}\fi
\ifx \citeauthoryear \undefined \def \citeauthoryear#1{#1}\fi
\ifx \endbibitem  \undefined \def \endbibitem {}\fi
\ifx \bconflocation  \undefined \def \bconflocation#1{#1}\fi
\ifx \arxivurl  \undefined \def \arxivurl#1{\textsf{#1}}\fi
\csname PreBibitemsHook\endcsname

\bibitem[\protect\citeauthoryear{Krizhevsky et~al.}{2012}]{krizhevsky2012imagenet}
\begin{botherref}
\oauthor{\bsnm{Krizhevsky}, \binits{A.}},
\oauthor{\bsnm{Sutskever}, \binits{I.}},
\oauthor{\bsnm{Hinton}, \binits{G.E.}}:
Imagenet classification with deep convolutional neural networks.
Advances in neural information processing systems
\textbf{25}
(2012)
\end{botherref}
\endbibitem

\bibitem[\protect\citeauthoryear{He et~al.}{2016}]{he2016deep}
\begin{bchapter}
\bauthor{\bsnm{He}, \binits{K.}},
\bauthor{\bsnm{Zhang}, \binits{X.}},
\bauthor{\bsnm{Ren}, \binits{S.}},
\bauthor{\bsnm{Sun}, \binits{J.}}:
\bctitle{Deep residual learning for image recognition}.
In: \bbtitle{Proceedings of the IEEE Conference on Computer Vision and Pattern Recognition},
pp. \bfpage{770}--\blpage{778}
(\byear{2016})
\end{bchapter}
\endbibitem

\bibitem[\protect\citeauthoryear{Niu et~al.}{2020}]{niu2020decade}
\begin{barticle}
\bauthor{\bsnm{Niu}, \binits{S.}},
\bauthor{\bsnm{Liu}, \binits{Y.}},
\bauthor{\bsnm{Wang}, \binits{J.}},
\bauthor{\bsnm{Song}, \binits{H.}}:
\batitle{A decade survey of transfer learning (2010--2020)}.
\bjtitle{IEEE Transactions on Artificial Intelligence}
\bvolume{1}(\bissue{2}),
\bfpage{151}--\blpage{166}
(\byear{2020})
\doiurl{10.1109/TAI.2021.3054609}
\end{barticle}
\endbibitem

\bibitem[\protect\citeauthoryear{Hosseinzadeh~Taher et~al.}{2021}]{hosseinzadeh2021systematic}
\begin{bchapter}
\bauthor{\bsnm{Hosseinzadeh~Taher}, \binits{M.R.}},
\bauthor{\bsnm{Haghighi}, \binits{F.}},
\bauthor{\bsnm{Feng}, \binits{R.}},
\bauthor{\bsnm{Gotway}, \binits{M.B.}},
\bauthor{\bsnm{Liang}, \binits{J.}}:
\bctitle{A systematic benchmarking analysis of transfer learning for medical image analysis}.
In: \bbtitle{Domain Adaptation and Representation Transfer, and Affordable Healthcare and AI for Resource Diverse Global Health: Third MICCAI Workshop, DART 2021, and First MICCAI Workshop, FAIR 2021, Held in Conjunction with MICCAI 2021, Strasbourg, France, September 27 and October 1, 2021, Proceedings 3},
pp. \bfpage{3}--\blpage{13}
(\byear{2021}).
\bcomment{Springer}
\end{bchapter}
\endbibitem

\bibitem[\protect\citeauthoryear{Neyshabur et~al.}{2020}]{neyshabur2020being}
\begin{barticle}
\bauthor{\bsnm{Neyshabur}, \binits{B.}},
\bauthor{\bsnm{Sedghi}, \binits{H.}},
\bauthor{\bsnm{Zhang}, \binits{C.}}:
\batitle{What is being transferred in transfer learning?}
\bjtitle{Advances in neural information processing systems}
\bvolume{33},
\bfpage{512}--\blpage{523}
(\byear{2020})
\end{barticle}
\endbibitem

\bibitem[\protect\citeauthoryear{Ericsson et~al.}{2022}]{ericsson2022self2}
\begin{barticle}
\bauthor{\bsnm{Ericsson}, \binits{L.}},
\bauthor{\bsnm{Gouk}, \binits{H.}},
\bauthor{\bsnm{Loy}, \binits{C.C.}},
\bauthor{\bsnm{Hospedales}, \binits{T.M.}}:
\batitle{Self-supervised representation learning: Introduction, advances, and challenges}.
\bjtitle{IEEE Signal Processing Magazine}
\bvolume{39}(\bissue{3}),
\bfpage{42}--\blpage{62}
(\byear{2022})
\doiurl{10.1109/MSP.2021.3134634}
\end{barticle}
\endbibitem

\bibitem[\protect\citeauthoryear{Russakovsky et~al.}{2015}]{russakovsky2015imagenet}
\begin{barticle}
\bauthor{\bsnm{Russakovsky}, \binits{O.}},
\bauthor{\bsnm{Deng}, \binits{J.}},
\bauthor{\bsnm{Su}, \binits{H.}},
\bauthor{\bsnm{Krause}, \binits{J.}},
\bauthor{\bsnm{Satheesh}, \binits{S.}},
\bauthor{\bsnm{Ma}, \binits{S.}},
\bauthor{\bsnm{Huang}, \binits{Z.}},
\bauthor{\bsnm{Karpathy}, \binits{A.}},
\bauthor{\bsnm{Khosla}, \binits{A.}},
\bauthor{\bsnm{Bernstein}, \binits{M.}}, \betal:
\batitle{Imagenet large scale visual recognition challenge}.
\bjtitle{International journal of computer vision}
\bvolume{115},
\bfpage{211}--\blpage{252}
(\byear{2015})
\doiurl{10.1007/s11263-015-0816-y}
\end{barticle}
\endbibitem

\bibitem[\protect\citeauthoryear{Raghu et~al.}{2019}]{raghu2019transfusion}
\begin{botherref}
\oauthor{\bsnm{Raghu}, \binits{M.}},
\oauthor{\bsnm{Zhang}, \binits{C.}},
\oauthor{\bsnm{Kleinberg}, \binits{J.}},
\oauthor{\bsnm{Bengio}, \binits{S.}}:
Transfusion: Understanding transfer learning for medical imaging.
Advances in neural information processing systems
\textbf{32}
(2019)
\end{botherref}
\endbibitem

\bibitem[\protect\citeauthoryear{Zhao et~al.}{2023}]{zhao2023comparison}
\begin{botherref}
\oauthor{\bsnm{Zhao}, \binits{Z.}},
\oauthor{\bsnm{Alzubaidi}, \binits{L.}},
\oauthor{\bsnm{Zhang}, \binits{J.}},
\oauthor{\bsnm{Duan}, \binits{Y.}},
\oauthor{\bsnm{Gu}, \binits{Y.}}:
A comparison review of transfer learning and self-supervised learning: Definitions, applications, advantages and limitations.
Expert Systems with Applications,
122807
(2023)
\doiurl{10.1016/j.eswa.2023.122807}
\end{botherref}
\endbibitem

\bibitem[\protect\citeauthoryear{Yang et~al.}{2020}]{yang2020transfer}
\begin{barticle}
\bauthor{\bsnm{Yang}, \binits{X.}},
\bauthor{\bsnm{He}, \binits{X.}},
\bauthor{\bsnm{Liang}, \binits{Y.}},
\bauthor{\bsnm{Yang}, \binits{Y.}},
\bauthor{\bsnm{Zhang}, \binits{S.}},
\bauthor{\bsnm{Xie}, \binits{P.}}:
\batitle{Transfer learning or self-supervised learning? a tale of two pretraining paradigms}.
\bjtitle{arXiv preprint arXiv:2007.04234}
(\byear{2020})
\doiurl{10.48550/arXiv.2007.04234}
\end{barticle}
\endbibitem

\bibitem[\protect\citeauthoryear{Azizi et~al.}{2023}]{azizi2023robust}
\begin{barticle}
\bauthor{\bsnm{Azizi}, \binits{S.}},
\bauthor{\bsnm{Culp}, \binits{L.}},
\bauthor{\bsnm{Freyberg}, \binits{J.}},
\bauthor{\bsnm{Mustafa}, \binits{B.}},
\bauthor{\bsnm{Baur}, \binits{S.}},
\bauthor{\bsnm{Kornblith}, \binits{S.}},
\bauthor{\bsnm{Chen}, \binits{T.}},
\bauthor{\bsnm{Tomasev}, \binits{N.}},
\bauthor{\bsnm{Mitrović}, \binits{J.}},
\bauthor{\bsnm{Strachan}, \binits{P.}}, \betal:
\batitle{Robust and data-efficient generalization of self-supervised machine learning for diagnostic imaging}.
\bjtitle{Nature Biomedical Engineering}
(\byear{2023})
\doiurl{10.1038/s41551-023-01049-7}
\end{barticle}
\endbibitem

\bibitem[\protect\citeauthoryear{Wang}{2021}]{wang2021overview}
\begin{bchapter}
\bauthor{\bsnm{Wang}, \binits{K.}}:
\bctitle{An overview of deep learning based small sample medical imaging classification}.
In: \bbtitle{2021 International Conference on Signal Processing and Machine Learning (CONF-SPML)},
pp. \bfpage{278}--\blpage{281}
(\byear{2021}).
\doiurl{10.1109/CONF-SPML54095.2021.00060} .
\bcomment{IEEE}
\end{bchapter}
\endbibitem

\bibitem[\protect\citeauthoryear{Razzak et~al.}{2018}]{razzak2018deep}
\begin{botherref}
\oauthor{\bsnm{Razzak}, \binits{M.I.}},
\oauthor{\bsnm{Naz}, \binits{S.}},
\oauthor{\bsnm{Zaib}, \binits{A.}}:
Deep learning for medical image processing: Overview, challenges and the future.
Classification in BioApps: Automation of Decision Making,
323--350
(2018)
\end{botherref}
\endbibitem

\bibitem[\protect\citeauthoryear{Pan and Yang}{2010}]{pan2010survey}
\begin{barticle}
\bauthor{\bsnm{Pan}, \binits{S.J.}},
\bauthor{\bsnm{Yang}, \binits{Q.}}:
\batitle{A survey on transfer learning}.
\bjtitle{IEEE Transactions on knowledge and data engineering}
\bvolume{22}(\bissue{10}),
\bfpage{1345}--\blpage{1359}
(\byear{2010})
\doiurl{10.1109/TKDE.2009.191}
\end{barticle}
\endbibitem

\bibitem[\protect\citeauthoryear{Kornblith et~al.}{2019}]{kornblith2019better}
\begin{bchapter}
\bauthor{\bsnm{Kornblith}, \binits{S.}},
\bauthor{\bsnm{Shlens}, \binits{J.}},
\bauthor{\bsnm{Le}, \binits{Q.V.}}:
\bctitle{Do better imagenet models transfer better?}
In: \bbtitle{Proceedings of the IEEE/CVF Conference on Computer Vision and Pattern Recognition},
pp. \bfpage{2661}--\blpage{2671}
(\byear{2019})
\end{bchapter}
\endbibitem

\bibitem[\protect\citeauthoryear{Zhang et~al.}{2022}]{zhang2022survey}
\begin{barticle}
\bauthor{\bsnm{Zhang}, \binits{W.}},
\bauthor{\bsnm{Deng}, \binits{L.}},
\bauthor{\bsnm{Zhang}, \binits{L.}},
\bauthor{\bsnm{Wu}, \binits{D.}}:
\batitle{A survey on negative transfer}.
\bjtitle{IEEE/CAA Journal of Automatica Sinica}
\bvolume{10}(\bissue{2}),
\bfpage{305}--\blpage{329}
(\byear{2022})
\doiurl{10.1109/JAS.2022.106004}
\end{barticle}
\endbibitem

\bibitem[\protect\citeauthoryear{Kolesnikov et~al.}{2020}]{kolesnikov2020big}
\begin{bchapter}
\bauthor{\bsnm{Kolesnikov}, \binits{A.}},
\bauthor{\bsnm{Beyer}, \binits{L.}},
\bauthor{\bsnm{Zhai}, \binits{X.}},
\bauthor{\bsnm{Puigcerver}, \binits{J.}},
\bauthor{\bsnm{Yung}, \binits{J.}},
\bauthor{\bsnm{Gelly}, \binits{S.}},
\bauthor{\bsnm{Houlsby}, \binits{N.}}:
\bctitle{Big transfer (bit): General visual representation learning}.
In: \bbtitle{Computer Vision--ECCV 2020: 16th European Conference, Glasgow, UK, August 23--28, 2020, Proceedings, Part V 16},
pp. \bfpage{491}--\blpage{507}
(\byear{2020}).
\bcomment{Springer}
\end{bchapter}
\endbibitem

\bibitem[\protect\citeauthoryear{Huh et~al.}{2016}]{huh2016makes}
\begin{barticle}
\bauthor{\bsnm{Huh}, \binits{M.}},
\bauthor{\bsnm{Agrawal}, \binits{P.}},
\bauthor{\bsnm{Efros}, \binits{A.A.}}:
\batitle{What makes imagenet good for transfer learning?}
\bjtitle{arXiv preprint arXiv:1608.08614}
(\byear{2016})
\doiurl{10.48550/arXiv.1608.08614}
\end{barticle}
\endbibitem

\bibitem[\protect\citeauthoryear{Morid et~al.}{2021}]{morid2021scoping}
\begin{barticle}
\bauthor{\bsnm{Morid}, \binits{M.A.}},
\bauthor{\bsnm{Borjali}, \binits{A.}},
\bauthor{\bsnm{Del~Fiol}, \binits{G.}}:
\batitle{A scoping review of transfer learning research on medical image analysis using imagenet}.
\bjtitle{Computers in biology and medicine}
\bvolume{128},
\bfpage{104115}
(\byear{2021})
\doiurl{10.1016/j.compbiomed.2020.104115}
\end{barticle}
\endbibitem

\bibitem[\protect\citeauthoryear{Patil and Sharma}{2024}]{patil2024automatic}
\begin{botherref}
\oauthor{\bsnm{Patil}, \binits{R.}},
\oauthor{\bsnm{Sharma}, \binits{S.}}:
Automatic glaucoma detection from fundus images using transfer learning.
Multimedia Tools and Applications,
1--20
(2024)
\doiurl{10.1007/s11042-024-18242-8}
\end{botherref}
\endbibitem

\bibitem[\protect\citeauthoryear{Alzubaidi et~al.}{2020}]{alzubaidi2020towards}
\begin{barticle}
\bauthor{\bsnm{Alzubaidi}, \binits{L.}},
\bauthor{\bsnm{Fadhel}, \binits{M.A.}},
\bauthor{\bsnm{Al-Shamma}, \binits{O.}},
\bauthor{\bsnm{Zhang}, \binits{J.}},
\bauthor{\bsnm{Santamar{\'\i}a}, \binits{J.}},
\bauthor{\bsnm{Duan}, \binits{Y.}},
\bauthor{\bsnm{R.~Oleiwi}, \binits{S.}}:
\batitle{Towards a better understanding of transfer learning for medical imaging: a case study}.
\bjtitle{Applied Sciences}
\bvolume{10}(\bissue{13}),
\bfpage{4523}
(\byear{2020})
\doiurl{10.3390/app10134523}
\end{barticle}
\endbibitem

\bibitem[\protect\citeauthoryear{Shurrab and Duwairi}{2022}]{shurrab2022self}
\begin{barticle}
\bauthor{\bsnm{Shurrab}, \binits{S.}},
\bauthor{\bsnm{Duwairi}, \binits{R.}}:
\batitle{Self-supervised learning methods and applications in medical imaging analysis: A survey}.
\bjtitle{PeerJ Computer Science}
\bvolume{8},
\bfpage{1045}
(\byear{2022})
\doiurl{10.7717/peerj-cs.1045}
\end{barticle}
\endbibitem

\bibitem[\protect\citeauthoryear{Ericsson et~al.}{2021}]{ericsson2022self}
\begin{barticle}
\bauthor{\bsnm{Ericsson}, \binits{L.}},
\bauthor{\bsnm{Gouk}, \binits{H.}},
\bauthor{\bsnm{Hospedales}, \binits{T.M.}}:
\batitle{Why do self-supervised models transfer? investigating the impact of invariance on downstream tasks}.
\bjtitle{arXiv preprint arXiv:2111.11398}
(\byear{2021})
\doiurl{10.48550/arXiv.2111.11398}
\end{barticle}
\endbibitem

\bibitem[\protect\citeauthoryear{Truong et~al.}{2021}]{truong2021transferable}
\begin{bchapter}
\bauthor{\bsnm{Truong}, \binits{T.}},
\bauthor{\bsnm{Mohammadi}, \binits{S.}},
\bauthor{\bsnm{Lenga}, \binits{M.}}:
\bctitle{How transferable are self-supervised features in medical image classification tasks?}
In: \bbtitle{Machine Learning for Health},
pp. \bfpage{54}--\blpage{74}
(\byear{2021}).
\bcomment{PMLR}
\end{bchapter}
\endbibitem

\bibitem[\protect\citeauthoryear{Del~Pup and Atzori}{2023}]{del2023applications}
\begin{barticle}
\bauthor{\bsnm{Del~Pup}, \binits{F.}},
\bauthor{\bsnm{Atzori}, \binits{M.}}:
\batitle{Applications of self-supervised learning to biomedical signals: a survey}.
\bjtitle{IEEE Access}
(\byear{2023})
\doiurl{10.1109/ACCESS.2023.3344531}
\end{barticle}
\endbibitem

\bibitem[\protect\citeauthoryear{Wu et~al.}{2023}]{wu2023self}
\begin{barticle}
\bauthor{\bsnm{Wu}, \binits{R.}},
\bauthor{\bsnm{Liang}, \binits{C.}},
\bauthor{\bsnm{Li}, \binits{Y.}},
\bauthor{\bsnm{Shi}, \binits{X.}},
\bauthor{\bsnm{Zhang}, \binits{J.}},
\bauthor{\bsnm{Huang}, \binits{H.}}:
\batitle{Self-supervised transfer learning framework driven by visual attention for benign--malignant lung nodule classification on chest ct}.
\bjtitle{Expert Systems with Applications}
\bvolume{215},
\bfpage{119339}
(\byear{2023})
\doiurl{10.1016/j.eswa.2022.119339}
\end{barticle}
\endbibitem

\bibitem[\protect\citeauthoryear{Romero et~al.}{2020}]{romero2020targeted}
\begin{barticle}
\bauthor{\bsnm{Romero}, \binits{M.}},
\bauthor{\bsnm{Interian}, \binits{Y.}},
\bauthor{\bsnm{Solberg}, \binits{T.}},
\bauthor{\bsnm{Valdes}, \binits{G.}}:
\batitle{Targeted transfer learning to improve performance in small medical physics datasets}.
\bjtitle{Medical physics}
\bvolume{47}(\bissue{12}),
\bfpage{6246}--\blpage{6256}
(\byear{2020})
\doiurl{10.1002/mp.14507}
\end{barticle}
\endbibitem

\bibitem[\protect\citeauthoryear{Alzubaidi et~al.}{2021}]{alzubaidi2021novel}
\begin{barticle}
\bauthor{\bsnm{Alzubaidi}, \binits{L.}},
\bauthor{\bsnm{Al-Amidie}, \binits{M.}},
\bauthor{\bsnm{Al-Asadi}, \binits{A.}},
\bauthor{\bsnm{Humaidi}, \binits{A.J.}},
\bauthor{\bsnm{Al-Shamma}, \binits{O.}},
\bauthor{\bsnm{Fadhel}, \binits{M.A.}},
\bauthor{\bsnm{Zhang}, \binits{J.}},
\bauthor{\bsnm{Santamar{\'\i}a}, \binits{J.}},
\bauthor{\bsnm{Duan}, \binits{Y.}}:
\batitle{Novel transfer learning approach for medical imaging with limited labeled data}.
\bjtitle{Cancers}
\bvolume{13}(\bissue{7}),
\bfpage{1590}
(\byear{2021})
\doiurl{10.3390/cancers13071590}
\end{barticle}
\endbibitem

\bibitem[\protect\citeauthoryear{Atasever et~al.}{2023}]{atasever2023comprehensive}
\begin{barticle}
\bauthor{\bsnm{Atasever}, \binits{S.}},
\bauthor{\bsnm{Azginoglu}, \binits{N.}},
\bauthor{\bsnm{Terzi}, \binits{D.S.}},
\bauthor{\bsnm{Terzi}, \binits{R.}}:
\batitle{A comprehensive survey of deep learning research on medical image analysis with focus on transfer learning}.
\bjtitle{Clinical Imaging}
\bvolume{94},
\bfpage{18}--\blpage{41}
(\byear{2023})
\doiurl{10.1016/j.clinimag.2022.11.003}
\end{barticle}
\endbibitem

\bibitem[\protect\citeauthoryear{Tan et~al.}{2024}]{tan2024self}
\begin{barticle}
\bauthor{\bsnm{Tan}, \binits{Z.}},
\bauthor{\bsnm{Yu}, \binits{Y.}},
\bauthor{\bsnm{Meng}, \binits{J.}},
\bauthor{\bsnm{Liu}, \binits{S.}},
\bauthor{\bsnm{Li}, \binits{W.}}:
\batitle{Self-supervised learning with self-distillation on covid-19 medical image classification}.
\bjtitle{Computer Methods and Programs in Biomedicine}
\bvolume{243},
\bfpage{107876}
(\byear{2024})
\doiurl{10.1016/j.cmpb.2023.107876}
\end{barticle}
\endbibitem

\bibitem[\protect\citeauthoryear{Muljo et~al.}{2023}]{muljo2023handling}
\begin{barticle}
\bauthor{\bsnm{Muljo}, \binits{H.H.}},
\bauthor{\bsnm{Pardamean}, \binits{B.}},
\bauthor{\bsnm{Elwirehardja}, \binits{G.N.}},
\bauthor{\bsnm{Hidayat}, \binits{A.A.}},
\bauthor{\bsnm{Sudigyo}, \binits{D.}},
\bauthor{\bsnm{Rahutomo}, \binits{R.}},
\bauthor{\bsnm{Cenggoro}, \binits{T.W.}}:
\batitle{Handling severe data imbalance in chest x-ray image classification with transfer learning using swav self-supervised pre-training}.
\bjtitle{Commun. Math. Biol. Neurosci.}
\bvolume{2023},
(\byear{2023})
\end{barticle}
\endbibitem

\bibitem[\protect\citeauthoryear{Zhang et~al.}{2023}]{zhang2023dive}
\begin{barticle}
\bauthor{\bsnm{Zhang}, \binits{C.}},
\bauthor{\bsnm{Zheng}, \binits{H.}},
\bauthor{\bsnm{Gu}, \binits{Y.}}:
\batitle{Dive into the details of self-supervised learning for medical image analysis}.
\bjtitle{Medical Image Analysis}
\bvolume{89},
\bfpage{102879}
(\byear{2023})
\doiurl{10.1016/j.media.2023.102879}
\end{barticle}
\endbibitem

\bibitem[\protect\citeauthoryear{Qayyum et~al.}{2020}]{qayyum2020secure}
\begin{barticle}
\bauthor{\bsnm{Qayyum}, \binits{A.}},
\bauthor{\bsnm{Qadir}, \binits{J.}},
\bauthor{\bsnm{Bilal}, \binits{M.}},
\bauthor{\bsnm{Al-Fuqaha}, \binits{A.}}:
\batitle{Secure and robust machine learning for healthcare: A survey}.
\bjtitle{IEEE Reviews in Biomedical Engineering}
\bvolume{14},
\bfpage{156}--\blpage{180}
(\byear{2020})
\doiurl{10.1109/RBME.2020.3013489}
\end{barticle}
\endbibitem

\bibitem[\protect\citeauthoryear{Schiappa et~al.}{2023}]{schiappa2023self}
\begin{barticle}
\bauthor{\bsnm{Schiappa}, \binits{M.C.}},
\bauthor{\bsnm{Rawat}, \binits{Y.S.}},
\bauthor{\bsnm{Shah}, \binits{M.}}:
\batitle{Self-supervised learning for videos: A survey}.
\bjtitle{ACM Computing Surveys}
\bvolume{55}(\bissue{13s}),
\bfpage{1}--\blpage{37}
(\byear{2023})
\doiurl{10.1145/3577925}
\end{barticle}
\endbibitem

\bibitem[\protect\citeauthoryear{Chen et~al.}{2020}]{chen2020simple}
\begin{bchapter}
\bauthor{\bsnm{Chen}, \binits{T.}},
\bauthor{\bsnm{Kornblith}, \binits{S.}},
\bauthor{\bsnm{Norouzi}, \binits{M.}},
\bauthor{\bsnm{Hinton}, \binits{G.}}:
\bctitle{A simple framework for contrastive learning of visual representations}.
In: \bbtitle{International Conference on Machine Learning},
pp. \bfpage{1597}--\blpage{1607}
(\byear{2020}).
\bcomment{PMLR}
\end{bchapter}
\endbibitem

\bibitem[\protect\citeauthoryear{Szegedy et~al.}{2017}]{szegedy2017inception}
\begin{bchapter}
\bauthor{\bsnm{Szegedy}, \binits{C.}},
\bauthor{\bsnm{Ioffe}, \binits{S.}},
\bauthor{\bsnm{Vanhoucke}, \binits{V.}},
\bauthor{\bsnm{Alemi}, \binits{A.}}:
\bctitle{Inception-v4, inception-resnet and the impact of residual connections on learning}.
In: \bbtitle{Proceedings of the AAAI Conference on Artificial Intelligence},
vol. \bseriesno{31-1},
pp. \bfpage{4278}--\blpage{4284}
(\byear{2017}).
\doiurl{10.1609/aaai.v31i1.11231}
\end{bchapter}
\endbibitem

\bibitem[\protect\citeauthoryear{Chollet}{2017}]{chollet2017xception}
\begin{bchapter}
\bauthor{\bsnm{Chollet}, \binits{F.}}:
\bctitle{Xception: Deep learning with depthwise separable convolutions}.
In: \bbtitle{Proceedings of the IEEE Conference on Computer Vision and Pattern Recognition},
pp. \bfpage{1251}--\blpage{1258}
(\byear{2017})
\end{bchapter}
\endbibitem

\bibitem[\protect\citeauthoryear{Howard et~al.}{2019}]{howard2019searching}
\begin{bchapter}
\bauthor{\bsnm{Howard}, \binits{A.}},
\bauthor{\bsnm{Sandler}, \binits{M.}},
\bauthor{\bsnm{Chu}, \binits{G.}},
\bauthor{\bsnm{Chen}, \binits{L.-C.}},
\bauthor{\bsnm{Chen}, \binits{B.}},
\bauthor{\bsnm{Tan}, \binits{M.}},
\bauthor{\bsnm{Wang}, \binits{W.}},
\bauthor{\bsnm{Zhu}, \binits{Y.}},
\bauthor{\bsnm{Pang}, \binits{R.}},
\bauthor{\bsnm{Vasudevan}, \binits{V.}}, \betal:
\bctitle{Searching for mobilenetv3}.
In: \bbtitle{Proceedings of the IEEE/CVF International Conference on Computer Vision},
pp. \bfpage{1314}--\blpage{1324}
(\byear{2019})
\end{bchapter}
\endbibitem

\bibitem[\protect\citeauthoryear{Al-Dhabyani et~al.}{2020}]{al2020dataset}
\begin{barticle}
\bauthor{\bsnm{Al-Dhabyani}, \binits{W.}},
\bauthor{\bsnm{Gomaa}, \binits{M.}},
\bauthor{\bsnm{Khaled}, \binits{H.}},
\bauthor{\bsnm{Fahmy}, \binits{A.}}:
\batitle{Dataset of breast ultrasound images}.
\bjtitle{Data in brief}
\bvolume{28},
\bfpage{104863}
(\byear{2020})
\doiurl{10.1016/j.dib.2019.104863}
\end{barticle}
\endbibitem

\bibitem[\protect\citeauthoryear{Hany}{2020}]{ha2020chest}
\begin{botherref}
\oauthor{\bsnm{Hany}, \binits{M.}}:
Chest CT-scan images dataset
(2020).
\url{https://www.kaggle.com/datasets/mohamedhanyyy/chest-ctscan-images}
\end{botherref}
\endbibitem

\bibitem[\protect\citeauthoryear{Pogorelov et~al.}{2017}]{pogorelov2017kvasir}
\begin{bchapter}
\bauthor{\bsnm{Pogorelov}, \binits{K.}},
\bauthor{\bsnm{Randel}, \binits{K.R.}},
\bauthor{\bsnm{Griwodz}, \binits{C.}},
\bauthor{\bsnm{Eskeland}, \binits{S.L.}},
\bauthor{\bsnm{Lange}, \binits{T.}},
\bauthor{\bsnm{Johansen}, \binits{D.}},
\bauthor{\bsnm{Spampinato}, \binits{C.}},
\bauthor{\bsnm{Dang-Nguyen}, \binits{D.-T.}},
\bauthor{\bsnm{Lux}, \binits{M.}},
\bauthor{\bsnm{Schmidt}, \binits{P.T.}}, \betal:
\bctitle{Kvasir: A multi-class image dataset for computer aided gastrointestinal disease detection}.
In: \bbtitle{Proceedings of the 8th ACM on Multimedia Systems Conference},
pp. \bfpage{164}--\blpage{169}
(\byear{2017}).
\doiurl{10.1145/3083187.3083212}
\end{bchapter}
\endbibitem

\bibitem[\protect\citeauthoryear{Kiefer et~al.}{2023}]{kiefer2023automated}
\begin{bchapter}
\bauthor{\bsnm{Kiefer}, \binits{R.}},
\bauthor{\bsnm{Abid}, \binits{M.}},
\bauthor{\bsnm{Ardali}, \binits{M.R.}},
\bauthor{\bsnm{Steen}, \binits{J.}},
\bauthor{\bsnm{Amjadian}, \binits{E.}}:
\bctitle{Automated fundus image standardization using a dynamic global foreground threshold algorithm}.
In: \bbtitle{2023 8th International Conference on Image, Vision and Computing (ICIVC)},
pp. \bfpage{460}--\blpage{465}
(\byear{2023}).
\doiurl{10.1109/ICIVC58118.2023.10270429} .
\bcomment{IEEE}
\end{bchapter}
\endbibitem

\bibitem[\protect\citeauthoryear{Tan et~al.}{2017}]{tan2017distant}
\begin{bchapter}
\bauthor{\bsnm{Tan}, \binits{B.}},
\bauthor{\bsnm{Zhang}, \binits{Y.}},
\bauthor{\bsnm{Pan}, \binits{S.}},
\bauthor{\bsnm{Yang}, \binits{Q.}}:
\bctitle{Distant domain transfer learning}.
In: \bbtitle{Proceedings of the AAAI Conference on Artificial Intelligence},
vol. \bseriesno{31-1},
pp. \bfpage{2604}--\blpage{2610}
(\byear{2017})
\end{bchapter}
\endbibitem

\bibitem[\protect\citeauthoryear{Niu et~al.}{2021}]{niu2021distant}
\begin{barticle}
\bauthor{\bsnm{Niu}, \binits{S.}},
\bauthor{\bsnm{Liu}, \binits{M.}},
\bauthor{\bsnm{Liu}, \binits{Y.}},
\bauthor{\bsnm{Wang}, \binits{J.}},
\bauthor{\bsnm{Song}, \binits{H.}}:
\batitle{Distant domain transfer learning for medical imaging}.
\bjtitle{IEEE Journal of Biomedical and Health Informatics}
\bvolume{25}(\bissue{10}),
\bfpage{3784}--\blpage{3793}
(\byear{2021})
\doiurl{10.1109/JBHI.2021.3051470}
\end{barticle}
\endbibitem

\bibitem[\protect\citeauthoryear{Radford}{2015}]{radford2015unsupervised}
\begin{barticle}
\bauthor{\bsnm{Radford}, \binits{A.}}:
\batitle{Unsupervised representation learning with deep convolutional generative adversarial networks}.
\bjtitle{arXiv preprint arXiv:1511.06434}
(\byear{2015})
\doiurl{10.48550/arXiv.1511.06434}
\end{barticle}
\endbibitem

\bibitem[\protect\citeauthoryear{Anagha and Aishwarya}{2021}]{chou2021ultra}
\begin{botherref}
\oauthor{\bsnm{Anagha}, \binits{C.}},
\oauthor{\bsnm{Aishwarya}, \binits{K.}}:
PCOS detection using ultrasound images
(2021).
\url{https://www.kaggle.com/datasets/anaghachoudhari/pcos-detection-using-ultrasound-images}
\end{botherref}
\endbibitem

\bibitem[\protect\citeauthoryear{Islam et~al.}{2022}]{islam2022vision}
\begin{barticle}
\bauthor{\bsnm{Islam}, \binits{M.N.}},
\bauthor{\bsnm{Hasan}, \binits{M.}},
\bauthor{\bsnm{Hossain}, \binits{M.K.}},
\bauthor{\bsnm{Alam}, \binits{M.G.R.}},
\bauthor{\bsnm{Uddin}, \binits{M.Z.}},
\bauthor{\bsnm{Soylu}, \binits{A.}}:
\batitle{Vision transformer and explainable transfer learning models for auto detection of kidney cyst, stone and tumor from ct-radiography}.
\bjtitle{Scientific Reports}
\bvolume{12}(\bissue{1}),
\bfpage{11440}
(\byear{2022})
\doiurl{10.1038/s41598-022-15634-4}
\end{barticle}
\endbibitem

\bibitem[\protect\citeauthoryear{Jha et~al.}{2021}]{jha2021comprehensive}
\begin{barticle}
\bauthor{\bsnm{Jha}, \binits{D.}},
\bauthor{\bsnm{Ali}, \binits{S.}},
\bauthor{\bsnm{Hicks}, \binits{S.}},
\bauthor{\bsnm{Thambawita}, \binits{V.}},
\bauthor{\bsnm{Borgli}, \binits{H.}},
\bauthor{\bsnm{Smedsrud}, \binits{P.H.}},
\bauthor{\bsnm{Lange}, \binits{T.}},
\bauthor{\bsnm{Pogorelov}, \binits{K.}},
\bauthor{\bsnm{Wang}, \binits{X.}},
\bauthor{\bsnm{Harzig}, \binits{P.}}, \betal:
\batitle{A comprehensive analysis of classification methods in gastrointestinal endoscopy imaging}.
\bjtitle{Medical image analysis}
\bvolume{70},
\bfpage{102007}
(\byear{2021})
\doiurl{10.1016/j.media.2021.102007}
\end{barticle}
\endbibitem

\bibitem[\protect\citeauthoryear{Gheflati and Rivaz}{2022}]{gheflati2022vision}
\begin{bchapter}
\bauthor{\bsnm{Gheflati}, \binits{B.}},
\bauthor{\bsnm{Rivaz}, \binits{H.}}:
\bctitle{Vision transformers for classification of breast ultrasound images}.
In: \bbtitle{2022 44th Annual International Conference of the IEEE Engineering in Medicine \& Biology Society (EMBC)},
pp. \bfpage{480}--\blpage{483}
(\byear{2022}).
\doiurl{10.1109/EMBC48229.2022.9871809} .
\bcomment{IEEE}
\end{bchapter}
\endbibitem

\bibitem[\protect\citeauthoryear{Deb and Jha}{2023}]{deb2023breast}
\begin{barticle}
\bauthor{\bsnm{Deb}, \binits{S.D.}},
\bauthor{\bsnm{Jha}, \binits{R.K.}}:
\batitle{Breast ultrasound image classification using fuzzy-rank-based ensemble network}.
\bjtitle{Biomedical Signal Processing and Control}
\bvolume{85},
\bfpage{104871}
(\byear{2023})
\doiurl{10.1016/j.bspc.2023.104871}
\end{barticle}
\endbibitem

\bibitem[\protect\citeauthoryear{Yue and Li}{2024}]{yue2024medmamba}
\begin{barticle}
\bauthor{\bsnm{Yue}, \binits{Y.}},
\bauthor{\bsnm{Li}, \binits{Z.}}:
\batitle{Medmamba: Vision mamba for medical image classification}.
\bjtitle{arXiv preprint arXiv:2403.03849}
(\byear{2024})
\doiurl{https://arxiv.org/abs/2403.03849}
\end{barticle}
\endbibitem

\bibitem[\protect\citeauthoryear{Dadgar and Neshat}{2022}]{dadgar2022comparative}
\begin{bchapter}
\bauthor{\bsnm{Dadgar}, \binits{S.}},
\bauthor{\bsnm{Neshat}, \binits{M.}}:
\bctitle{Comparative hybrid deep convolutional learning framework with transfer learning for diagnosis of lung cancer}.
In: \bbtitle{International Conference on Soft Computing and Pattern Recognition},
pp. \bfpage{296}--\blpage{305}
(\byear{2022}).
\bcomment{Springer}
\end{bchapter}
\endbibitem

\bibitem[\protect\citeauthoryear{Mamun et~al.}{2023}]{mamun2023lcdctcnn}
\begin{bchapter}
\bauthor{\bsnm{Mamun}, \binits{M.}},
\bauthor{\bsnm{Mahmud}, \binits{M.I.}},
\bauthor{\bsnm{Meherin}, \binits{M.}},
\bauthor{\bsnm{Abdelgawad}, \binits{A.}}:
\bctitle{Lcdctcnn: Lung cancer diagnosis of ct scan images using cnn based model}.
In: \bbtitle{2023 10th International Conference on Signal Processing and Integrated Networks (SPIN)},
pp. \bfpage{205}--\blpage{212}
(\byear{2023}).
\doiurl{10.1109/SPIN57001.2023.10116075} .
\bcomment{IEEE}
\end{bchapter}
\endbibitem

\bibitem[\protect\citeauthoryear{Dong et~al.}{2023}]{dong2023local}
\begin{bchapter}
\bauthor{\bsnm{Dong}, \binits{Z.}},
\bauthor{\bsnm{Xu}, \binits{B.}},
\bauthor{\bsnm{Shi}, \binits{J.}},
\bauthor{\bsnm{Zheng}, \binits{L.}}:
\bctitle{Local and global feature interaction network for endoscope image classification}.
In: \bbtitle{International Conference on Image and Graphics},
pp. \bfpage{412}--\blpage{424}
(\byear{2023}).
\bcomment{Springer}
\end{bchapter}
\endbibitem

\bibitem[\protect\citeauthoryear{Mukhtorov et~al.}{2023}]{mukhtorov2023endoscopic}
\begin{barticle}
\bauthor{\bsnm{Mukhtorov}, \binits{D.}},
\bauthor{\bsnm{Rakhmonova}, \binits{M.}},
\bauthor{\bsnm{Muksimova}, \binits{S.}},
\bauthor{\bsnm{Cho}, \binits{Y.-I.}}:
\batitle{Endoscopic image classification based on explainable deep learning}.
\bjtitle{Sensors}
\bvolume{23}(\bissue{6}),
\bfpage{3176}
(\byear{2023})
\doiurl{10.3390/s23063176}
\end{barticle}
\endbibitem

\bibitem[\protect\citeauthoryear{Wang et~al.}{2023}]{wang2023vision}
\begin{barticle}
\bauthor{\bsnm{Wang}, \binits{W.}},
\bauthor{\bsnm{Yang}, \binits{X.}},
\bauthor{\bsnm{Tang}, \binits{J.}}:
\batitle{Vision transformer with hybrid shifted windows for gastrointestinal endoscopy image classification}.
\bjtitle{IEEE Transactions on Circuits and Systems for Video Technology}
(\byear{2023})
\doiurl{10.1109/TCSVT.2023.3277462}
\end{barticle}
\endbibitem

\bibitem[\protect\citeauthoryear{Patel et~al.}{2024}]{patel2024classification}
\begin{bchapter}
\bauthor{\bsnm{Patel}, \binits{V.}},
\bauthor{\bsnm{Patel}, \binits{K.}},
\bauthor{\bsnm{Goel}, \binits{P.}},
\bauthor{\bsnm{Shah}, \binits{M.}}:
\bctitle{Classification of gastrointestinal diseases from endoscopic images using convolutional neural network with transfer learning}.
In: \bbtitle{2024 5th International Conference on Intelligent Communication Technologies and Virtual Mobile Networks (ICICV)},
pp. \bfpage{504}--\blpage{508}
(\byear{2024}).
\doiurl{10.1109/ICICV62344.2024.00085} .
\bcomment{IEEE}
\end{bchapter}
\endbibitem

\bibitem[\protect\citeauthoryear{Albahri et~al.}{2023}]{albahri2023systematic}
\begin{barticle}
\bauthor{\bsnm{Albahri}, \binits{A.}},
\bauthor{\bsnm{Duhaim}, \binits{A.M.}},
\bauthor{\bsnm{Fadhel}, \binits{M.A.}},
\bauthor{\bsnm{Alnoor}, \binits{A.}},
\bauthor{\bsnm{Baqer}, \binits{N.S.}},
\bauthor{\bsnm{Alzubaidi}, \binits{L.}},
\bauthor{\bsnm{Albahri}, \binits{O.}},
\bauthor{\bsnm{Alamoodi}, \binits{A.}},
\bauthor{\bsnm{Bai}, \binits{J.}},
\bauthor{\bsnm{Salhi}, \binits{A.}}, \betal:
\batitle{A systematic review of trustworthy and explainable artificial intelligence in healthcare: Assessment of quality, bias risk, and data fusion}.
\bjtitle{Information Fusion}
(\byear{2023})
\doiurl{10.1016/j.inffus.2023.03.008}
\end{barticle}
\endbibitem

\bibitem[\protect\citeauthoryear{Quinn et~al.}{2022}]{quinn2022three}
\begin{barticle}
\bauthor{\bsnm{Quinn}, \binits{T.P.}},
\bauthor{\bsnm{Jacobs}, \binits{S.}},
\bauthor{\bsnm{Senadeera}, \binits{M.}},
\bauthor{\bsnm{Le}, \binits{V.}},
\bauthor{\bsnm{Coghlan}, \binits{S.}}:
\batitle{The three ghosts of medical ai: Can the black-box present deliver?}
\bjtitle{Artificial intelligence in medicine}
\bvolume{124},
\bfpage{102158}
(\byear{2022})
\doiurl{10.1016/j.artmed.2021.102158}
\end{barticle}
\endbibitem

\bibitem[\protect\citeauthoryear{Zhou et~al.}{2016}]{zhou2016learning}
\begin{bchapter}
\bauthor{\bsnm{Zhou}, \binits{B.}},
\bauthor{\bsnm{Khosla}, \binits{A.}},
\bauthor{\bsnm{Lapedriza}, \binits{A.}},
\bauthor{\bsnm{Oliva}, \binits{A.}},
\bauthor{\bsnm{Torralba}, \binits{A.}}:
\bctitle{Learning deep features for discriminative localization}.
In: \bbtitle{Proceedings of the IEEE Conference on Computer Vision and Pattern Recognition},
pp. \bfpage{2921}--\blpage{2929}
(\byear{2016})
\end{bchapter}
\endbibitem

\bibitem[\protect\citeauthoryear{Selvaraju et~al.}{2017}]{selvaraju2017grad}
\begin{bchapter}
\bauthor{\bsnm{Selvaraju}, \binits{R.R.}},
\bauthor{\bsnm{Cogswell}, \binits{M.}},
\bauthor{\bsnm{Das}, \binits{A.}},
\bauthor{\bsnm{Vedantam}, \binits{R.}},
\bauthor{\bsnm{Parikh}, \binits{D.}},
\bauthor{\bsnm{Batra}, \binits{D.}}:
\bctitle{Grad-cam: Visual explanations from deep networks via gradient-based localization}.
In: \bbtitle{Proceedings of the IEEE International Conference on Computer Vision},
pp. \bfpage{618}--\blpage{626}
(\byear{2017})
\end{bchapter}
\endbibitem

\end{thebibliography}



\end{document}